%% file: iclr2020_conference.tex
\documentclass{article} % For LaTeX2e
\usepackage{iclr2020_conference,times}

\input{math_commands.tex}

\usepackage{hyperref}
\usepackage{url}

\usepackage{times}
\usepackage{latexsym}

\newcommand*{\affaddr}[1]{#1} % No op here. Customize it for different styles.
\newcommand*{\affmark}[1][*]{\textsuperscript{#1}}

\usepackage{subcaption}
\usepackage{amsmath}
\usepackage{pgfplots}
\usepackage{tikz}
\usepackage{multirow}

\usepackage{arydshln}

\usepackage{algorithm}
\usepackage[noend]{algpseudocode}
\usepackage{mathtools}
\usepackage{amssymb}

\include{plots}
\definecolor{bblue}{HTML}{4F81BD}
\definecolor{rred}{HTML}{C0504D}
\definecolor{ggreen}{HTML}{9BBB59}
\definecolor{ppurple}{HTML}{9F4C7C}
\definecolor{Dark scarlet}{HTML}{560319}
\definecolor{Forest green}{HTML}{1E4D2B}

\title{%Multi-view and Multi-source Transfers in Neural Topic Modeling\\
Multi-view and Multi-source Transfers in Neural Topic Modeling with Pretrained Topic and Word Embeddings% via Topic and Word Pools%TopicPool and WordPool
}

\iclrfinalcopy
\author{Pankaj Gupta\affmark[1,2], Yatin Chaudhary\affmark[1], Hinrich Sch\"{u}tze\affmark[2]\\ 
 \affaddr{\affmark[1]Corporate Technology, Machine-Intelligence (MIC-DE), Siemens AG  Munich, Germany}\\
  \affaddr{\affmark[2]CIS, University of Munich (LMU) Munich, Germany} \\
  {\tt \{pankaj.gupta, yatin.chaudhary\}@siemens.com}\\
}

\begin{document}
\maketitle

\begin{abstract}

%Word embeddings are learned from local context windows, whereas topic models take a more global view. 
Though word embeddings and topics are complementary representations, 
% in the sense that how they learn from word occurrences in a text corpora, 
several past works have only used pre-trained word embeddings in (neural) topic modeling to address data sparsity problem in short text or small collection of documents.
However, no prior work has employed (pre-trained latent) topics in transfer learning paradigm. 
%Essentially, word embeddings are fine-granularity (i.e., {\it local view}) representations in contrast to coarse-granularity (i.e., global view) in latent topics,  
In this paper, we propose an approach to 
%(1) perform knowledge transfer from a {\it TopicPool} of latent topics obtained from a large source corpus, and 
(1) perform knowledge transfer using latent topics obtained from a large source corpus, and 
(2) jointly transfer knowledge via the two representations (or views)
in neural topic modeling to improve topic quality, better deal with polysemy and data sparsity issues in a target corpus. 
In doing so, we first accumulate topics and word representations from one or many source corpora to build a pool of topics 
%(i.e., {\it TopicPool}) 
and word vectors.  
%(i.e., {\it WordPool}).  
%Moreover, 
Then, we identify one or multiple relevant source domain(s) and take advantage of corresponding topics and word features via 
the respective pools to guide meaningful learning in the sparse target domain. 
%Within neural topic modeling, 
We quantify the quality of topic and document representations via generalization (perplexity), %on unseen documents, 
interpretability (topic coherence) and information retrieval (IR) using short-text, long-text, small and large document collections from news and medical domains. 
We have demonstrated the state-of-the-art results on topic modeling with the proposed framework.  
% in unsupervised knowledge transfer within neural topic modeling that jointly exploits word and  topically contextualized features from several domains.
% i.e., {\it multi-view} from one or many related or distant domains, i.e., {\it multi-source}.  
%To achieve it, %in the sparse settings of the target domain, 
%we guide the generative process of learning the hidden topics in the target domain by latent topic vectors (i.e., coarse-grained) from a source domain(s) such that the hidden topics in the target domain get more meaningful and representative in explaining the target corpus. We quantify the quality of hidden topics via generalization (perplexity) on unseen documents, interpretability (topic coherence),  text retrieval and classification. 

\end{abstract}

\section{Introduction}\label{sec:introduction}

Probabilistic topic models, such as LDA \citep{DBLP:journals/jmlr/BleiNJ03}, Replicated Softmax (RSM) \citep{DBLP:conf/nips/SalakhutdinovH09} and Document Neural Autoregressive Distribution Estimator (DocNADE) \citep{DBLP:conf/nips/LarochelleL12} 
%,  DBLP:journals/jmlr/LaulyZAL17} 
are often used to extract topics from text collections and learn latent document representations %that can be used 
to perform natural language processing tasks, such as information retrieval (IR).
%document classification, etc. 
Though they have been shown to be powerful in modeling large text corpora,  
the topic modeling (TM) still remains challenging especially in the sparse-data setting,
especially for the cases where word co-occurrence data is insufficient 
e.g., %in settings with a small number of word occurrences for instance in short text or data sparsity in a corpus of few documents. 
on short text or a corpus of few documents. % due to a small number of word occurrences. 
% is difficult due to limited context or small number of word co-occurrences. 
To this end, several works \citep{P15-1077, DBLP:journals/tacl/NguyenBDJ15, pankajgupta:2019iDocNADEe} have introduced external knowledge in traditional topic models via word embeddings \citet{D14-1162}. 
However, no prior work in topic modeling has employed topical embeddings (obtained from large document collection(s)),   complementary to word embeddings. % in order to learn quality topics and improve predictive performance for a sparse. 

\iffalse
Several attempts have been made in order to handle the data sparsity issues in topic models, such as \citet{DBLP:conf/nips/PettersonSCBN10} introduced word similarity via thesauri and dictionaries,   %\citep{cite} leveraged web search results to augment inofrmation in the short texts 
and \citet{P15-1077, DBLP:journals/tacl/NguyenBDJ15} used word embeddings into LDA based models. Recently, \citet{pankajgupta:2019iDocNADEe} extends DocNADE by employing the pre-trained word embeddings from \citet{D14-1162} in a neural network based topic learning. 
\fi
%The increasing volume of short texts generated on social media sites, such as Twitter or Facebook, creates a great demand for effective and efficient topic modeling approaches
\iffalse
In context of topic modeling  \citep{DBLP:journals/jmlr/BleiNJ03}, a topic is defined as as the multinomial distribution of over vocabulary of words in the corpus and a topic mixture abstractly describes the underlying semantic structure of a document. 
Importantly in topic models,  a topic assigned to a given word occurrence equally depends on all the other words appearing in the same document and therefore, a topic learns word occurrences across documents and encodes a {\it coarse-grained} description of the corpus. Moreover, the thematic structures underlying in the corpus is described by a mixture of several {\it coarse-grained} descriptions, i.e., topics. 
In terms of interpretability, a topic semantic is defined by the high probability words in a distribution over a vocabulary. 
\fi

{\it Local vs Global Views}: 
%While the %two representations: 
Though word embeddings \citep{D14-1162} and topics are complementary in how they represent the meaning, they are distinctive in how they learn from word occurrences observed in text corpora.  Word embeddings have {\it local} context ({\it view}) in the sense that they are learned based on local collocation pattern in a text corpus, where the representation of each word either depends on a local context window %or where the context is limited by a local context window 
\citep{DBLP:conf/nips/MikolovSCCD13} or is a function of its sentence(s) \citep{N18-1202}. 
Consequently, the word occurrences are modeled in a {\it fine-granularity}. 
% and  
%the word embeddings can not capture the thematic structures or topical semantics in the underlying corpus. 
%In other words, the word embedding vectors encode local dependencies and thus, the word embeddings are referred as the {\it fine-grained} representation in context of this work.
On other hand, a topic \citep{DBLP:journals/jmlr/BleiNJ03} has a {\it global} word context ({\it view}): 
%in the sense that 
%a topic model 
TM infers topic distributions across documents in the corpus 
and assigns a topic to each word occurrence,  
where the assignment is equally dependent on all other words appearing in the same document. 
Therefore, it learns from word occurrences across documents and encodes a {\it coarse-granularity} description. 
\iffalse
Specifically, 
%and are learned by accounting word occurrences and distributions across the corpus. Specifically, 
a latent topic \citep{DBLP:conf/sigir/ShiLJSL17} that is assigned to a given word occurrence equally depends on all the other words
 that occur in the same document and it learns word occurrences across documents and encodes a {\it coarse-granularity} description. % of the corpus. 
%take a more global view, looking at the word
%distributions across the corpus to assign a topic to each word occurrence. 
% in the sense that the topic which is assigned to a given word occurrence (in the case of LDA) equally depends on all the other words that appear in the same document
\fi
Unlike topics, the word embeddings can not capture the thematic structures (topical semantics) in the underlying corpus.

Consider the following topics ($Z_1$-$Z_4$), where ($Z_1$-$Z_3$) are respectively obtained from different (high-resource) source ($\mathcal{S}^1$-$\mathcal{S}^3$) domains whereas $Z_4$ from the (low-resource) target domain $\mathcal{T}$ in the data-sparsity setting:

$\qquad$ $Z_1$ ($\mathcal{S}^1$):  {\it profit, growth,  stocks, {\bf apple}, fall, consumer, buy, billion, shares} $\rightarrow$ {\it Trading}

$\qquad$ $Z_2$($\mathcal{S}^2$): {\it smartphone, ipad, {\bf apple}, app, iphone, devices, phone, tablet} $\rightarrow$ {\it Product Line}

$\qquad$ $Z_3$ ($\mathcal{S}^3$): {\it microsoft, mac, linux, ibm, ios, {\bf apple}, xp, windows} $\rightarrow$ {\it Operating System/Company}
 
%$T_4$ ($\mathcal{S}^4$): {\it {\bf apple}, power, treatment, diagnosed, update, computer, condition, test} $\rightarrow$ {\it Utility}
$\qquad$ $Z_4$ ($\mathcal{T}$): {\it {\bf apple}, talk, computers, shares,  disease,  driver, electronics, profit, ios} $\rightarrow$ $?$

Usually, top words associated with topics learned on a large corpus are semantically coherent, %and represent meaningful semantics, 
e.g., {\it Trading, Product Line}, etc.      
However in sparse-data setting, topics (e.g., $Z_4$) are %semantically 
incoherent ({{\it noisy}) 
%, e.g., %
and therefore, it is difficult to infer meaningful semantics. 
Additionally, notice that the word {\it apple} is topically/thematically contextualized (topic-word association) in different semantics in $\mathcal{S}^1$-$\mathcal{S}^3$ and referring to a {\it company}.

%Unlike the topics, the top-5 nearest neighbors (NN) of {\it apple} (below) in the embeddings \citep{DBLP:conf/nips/MikolovSCCD13} space suggest that it refers to a {\it fruit}. 
Unlike the topics, word embeddings encode syntactic and semantic relatedness in fine-granularity and therefore, do not capture thematic structures. 
For instance, the top-5 nearest neighbors (NN) of {\it apple} (below) in the embeddings \citep{DBLP:conf/nips/MikolovSCCD13} space suggest that it refers to a {\it fruit}; 
however, they do not express anything about its thematic context, e.g., {\it Health}.

{%\small 
$\qquad$ $\qquad$ $\qquad$ $\qquad$ {\bf apple} $\xRightarrow[]{NN}$ {\it apples, pear, fruit, berry, pears, strawberry} 

$\qquad$  $\qquad$ $\qquad$ $\qquad$ {\bf fall} $\xRightarrow[]{NN}$ {\it falling, falls, drop, tumble, rise, plummet, fell}
}

Similarly for the word {\it fall}, it is difficult to infer its coarse-grained description, e.g, {\it Trading} as expressed by the topic $Z_1$.  

\iffalse
While the word embeddings encode syntactic and semantic relatedness in its vectors based on the co-occurrence statistics in a context, the context is either limited by a window-size \citep{DBLP:conf/nips/MikolovSCCD13} or a function of its sentence \citep{N18-1202}. Consequently, the word occurrences are modeled in a fine granularity and the word embeddings learned do not capture the topical semantics at the document level (or thematic structures of the underling corpus). In other words, the word embedding vectors encode local dependencies and thus, the word embeddings are referred as the {\it fine-grained} representation in context of this work.

Example of {\it fine-granularity}: Top-5 Nearest Neighbors (NN) of {\it apple} and {\it fall} in the word embeddings \citep{DBLP:conf/nips/MikolovSCCD13} space: 

{\bf apple} $\xRightarrow[]{NN}$ {\it apples, pear, fruit, berry, pears}

{\bf fall} $\xRightarrow[]{NN}$ {\it falling, falls, drop, tumble, rise}
\fi

%table of notations 
\begin{table*}[t]
\center
\renewcommand*{\arraystretch}{1.2}
\resizebox{.995\textwidth}{!}{
\begin{tabular}{c|l||c|l}
%\hline
\multicolumn{1}{c|}{\bf Notation} & \multicolumn{1}{c||}{\bf Description} & \multicolumn{1}{c|}{\bf Notation} & \multicolumn{1}{c}{\bf Description} \\ \hline
LVT, GVT  &  Local-view Transfer, Global-view Transfer  					& ${\bf A}^{k} \in \mathbb{R}^{H \times H}$    & Topic-alignment in $\mathcal{T}$ and ${\bf Z}^{k}$   \\ 
MVT, MST &   Multi-view Transfer, Multi-source Transfer 		 			& $ K, D$  & 	Vocabulary size, document size			\\
$\mathcal{T}$, $\mathcal{S}$  & A target domain, a set of source domains  		& $E$,  $H$ & Word embedding dimension, \#topics \\
$ \lambda^{k}$ 	 & Degree of relevance of ${\bf E}^{k}$ in $\mathcal{T}$ 	& ${\bf b} \in \mathbb{R}^{K}$, ${\bf c} \in \mathbb{R}^{H}$   & Visible-bias, hidden-bias\\
$ \gamma^{k}$  &	Degree of imitation of ${\bf Z}^{k}$ by ${\bf W}$			& ${\bf v}$, $k$, $\mathcal{L}$  & An input document, $k$th source, loss	\\
${\bf E}^{k} \in \mathbb{R}^{E \times K}$,   & Word embeddings of $k$th source & ${\bf W} \in \mathbb{R}^{H \times K}$   & Encoding matrix of DocNADE  in $\mathcal{T}$\\ 
${\bf Z}^{k} \in \mathbb{R}^{H \times K}$   & Topic embeddings of $k$th source  & ${\bf U} \in \mathbb{R}^{K \times H}$   & Decoding matrix of DocNADE %\\ \hline	 
\end{tabular}}
\caption{Description of the notations used in this work}
\label{tab:notations}
\end{table*}

%About topical context of apple......and word embedding of apple....negative transfer....better dismbuituation....deal with polysemy..
%{\bf Motivation (1)}: 
%Past works 
{\bf Motivation (1) Knowledge transfer via Complementary Representations (both word and topic representations)}: 
%Recent works such as \citet{P15-1077}, \citet{DBLP:journals/tacl/NguyenBDJ15} and \citet{pankajgupta:2019iDocNADEe} 
%have shown that TM can be improved by using external knowledge, e.g., word embeddings especially for short text or small collections to alleviate sparsity issues. 
Essentially, the application of TM aims to discover hidden thematic structures (i.e., topics) in text collection; however, it is challenging in data sparsity settings, e.g, in a short and/or small collection.  
This leads to suboptimal text representations and incoherent topics (e.g., topic $Z_4$). 

To alleviate the data sparsity issues, recent works such as \citet{P15-1077}, \citet{DBLP:journals/tacl/NguyenBDJ15} and \citet{pankajgupta:2019iDocNADEe} 
have shown that TM can be improved by introducing external knowledge, where they leverage pre-trained word embeddings (i.e., local view) only. 
However, the word embeddings ignore the thematically contextualized structures (i.e., document-level semantics), and can not deal with ambiguity.
Given that the word and topic representations encode complementary information, {\bf no} prior work has considered knowledge transfer via (pre-trained latent) topics (i.e., global view) from a large corpora.  

%In this work, we demonstrate knowledge transfer in TM jointly using word and topic representations to deal with the data sparsity.  

{\bf Motivation (2) Knowledge transfer via multiple sources of word and topic representations}: 
Knowledge transfer via word embeddings is vulnerable to negative transfer  \citep{DBLP:conf/aaai/CaoPZYY10} on the target domain when domains are shifted and not handled properly. 
% leading to a negative transfer \citep{DBLP:conf/aaai/CaoPZYY10} on the target domain if they are not handled properly. 
%-- only LVT, no GVT, no MST, title as GVT....
%For example, consider a short-text document ${\bf v}$ in the target domain $\mathcal{T}$
%and the top-5 Nearest Neighbors (NN) of {\it apple} in the embeddings \citep{D14-1162} space,
%${\bf v}$: {\small \texttt{Apple gained its market share as iPhone sales grow}}
%${\bf v}$: {\small \texttt{Apple gained its US market shares}} 
For instance, consider a short-text document ${\bf v}$:  [{\texttt{apple gained its US market shares}}] in the target domain $\mathcal{T}$. 
Here%\footnote{TM ignores punctuation, capitalization, stop words, etc.}
, the word ${\it apple}$  refers to a {\it company},  and hence  
%In such case, 
the word vector of {\it apple}  (about {\it fruit}) is an irrelevant source of knowledge transfer for both ${\bf v}$ and the topic $Z_4$. 
In contrast, one can better model ${\bf v}$ and amend the noisy $Z_4$ for coherence, 
given the meaningful word and topic representations. % (e.g., $Z_1$-$Z_3$).%. via latent topic features.   

%Notice that the word vector of {\it apple} intuitively means a {\it  fruit} whereas {\it apple} refers to a {\it company} in ${\bf v}$.
%In such a scenarios, the external knowledge (i.e., word embeddings) is an irrelevant source in improving performance and representations on $\mathcal{T}$. Moreover, the word vector of {\it apple} will neither improve the noisy topic $T_5$ learned in $\mathcal{T}$. On other hand, one can better model ${\bf v}$ and amend $T_5$ for coherence, 
%given meaningful representations in form of latent topic features $T_1$-$T_4$ from a large corpus(s).    

%{\bf Motivation (2)}: 
%In topic modeling, 
Often, there are several topic-word associations in different domains, e.g., in topics $Z_1$-$Z_3$. 
Given a noisy topic $Z_4$ in $\mathcal{T}$ and meaningful topics $Z_1$-$Z_3$ of $\mathcal{S}^1$-$\mathcal{S}^3$, we %want to
identify multiple relevant (source) domains %where knowledge can be transferred, 
and advantageously transfer their word and topic representations in order to facilitate meaningful learning in the sparse corpus, $\mathcal{T}$. 

{\bf Contribution (1)} To our knowledge, it is the {\it first work} in unsupervised topic modeling framework that introduces (external) knowledge transfer via (a) {\it Global-view Transfer}: latent topic representations (thematically contextualized) instead of using word embeddings exclusively, and (b) {\it Multi-view Transfer}: jointly using both the word and topic representations from a large source corpus in order to deal with polysemy and alleviate data sparsity issues in a small target corpus.  

{\bf Contribution (2)} {\it Multi-source Transfer}: Moreover, we first learn word and topic representations on multiple source domains and then perform {\it multi-view} and {\it multi-source} 
knowledge transfers within neural topic modeling by jointly using the complementary representations. % to improve a sparse target domain.%{\it word embeddings} and {\it latent topic features} %(complementary to each other) 
%from one or {\it many} related or distant source domains.   
%To our knowledge, it is the first attempt in unsupervised knowledge transfer within neural topic modeling that jointly exploits word and  topically contextualized features from several domains.
% i.e., {\it multi-view} from one or many related or distant domains, i.e., {\it multi-source}.  
%To achieve it, %in the sparse settings of the target domain, 
In doing so, 
we guide the (unsupervised) generative process of learning hidden topics of the target domain by word and latent topic features 
from a source domain(s) such that the hidden topics on the target become meaningful. % and representative in explaining the target corpus. 

\iffalse
{\bf Contribution(s)}: To better deal with polysemy and alleviate data sparsity issues, % short text or small collection of documents,
we introduce an approach to transfer latent topic features (thematically contextualized) instead of using word embeddings exclusively.  

Moreover, we first learn word and topic representations on multiple source domains and then perform {\it multi-view} and {\it multi-source} 
knowledge transfers within neural topic modeling by jointly using the complementary representations. % to improve a sparse target domain.%{\it word embeddings} and {\it latent topic features} %(complementary to each other) 
%from one or {\it many} related or distant source domains.   
%To our knowledge, it is the first attempt in unsupervised knowledge transfer within neural topic modeling that jointly exploits word and  topically contextualized features from several domains.
% i.e., {\it multi-view} from one or many related or distant domains, i.e., {\it multi-source}.  
%To achieve it, %in the sparse settings of the target domain, 
To do so, 
we guide the (unsupervised) generative process of learning hidden topics of the target domain by word and latent topic features 
from a source domain(s) such that the hidden topics on the target become meaningful. % and representative in explaining the target corpus. 

To our knowledge, it is the first unsupervised neural topic modeling framework that jointly leverages (external) complementary knowledge: latent word and topic features 
from one or many sources to alleviate data-sparsity issues.  
\fi

We evaluate the effectiveness of our transfer learning approaches in neural topic modeling  %using short-text, long-text, small and large document collections from news and medical domains. 
% approaches 
using 7 (5 low-resource and 2 high-resource) target and 5 (high-resource) source corpora from news and medical domains, consisting of  short-text, long-text, small and large document collections. 
Particularly, we quantify the quality of text representations via generalization (perplexity), interpretability (topic coherence) and text retrieval. 
%using short-text, long-text, small and large document collections from news and medical domains. 
%for an unsupervised knowledge transfer mechanism for neural topic modeling that jointly exploits word and topic features from several sources.
{\it The code is available in supplementary}.

\begin{figure}[t]
  \centering
  \includegraphics[scale=.67]{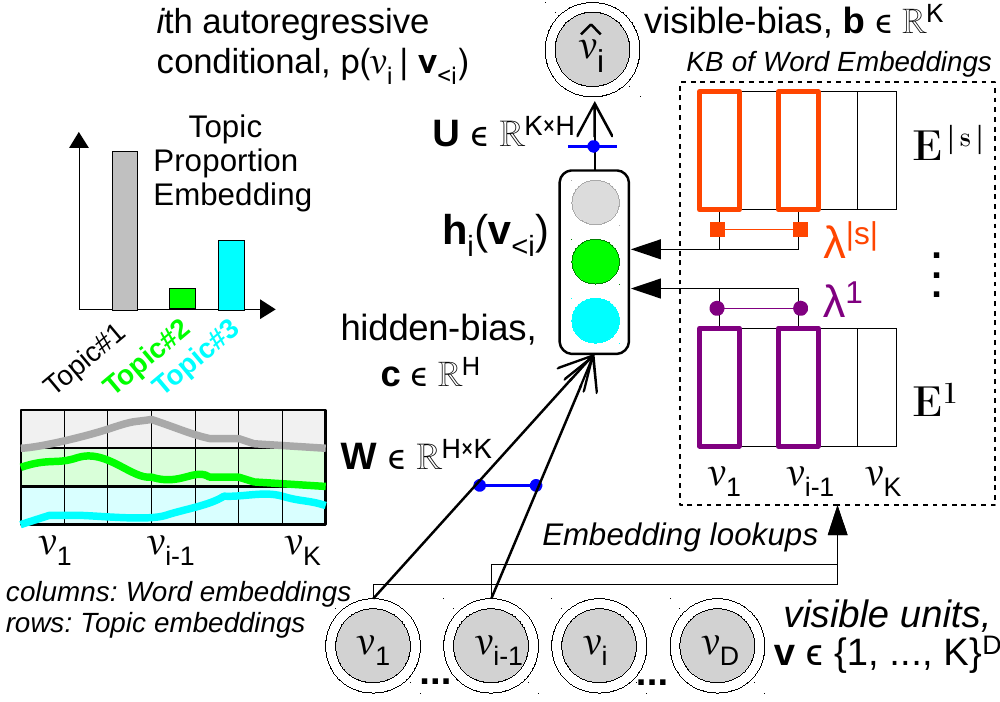} \ \ \ \ \ 
\includegraphics[scale=.65]{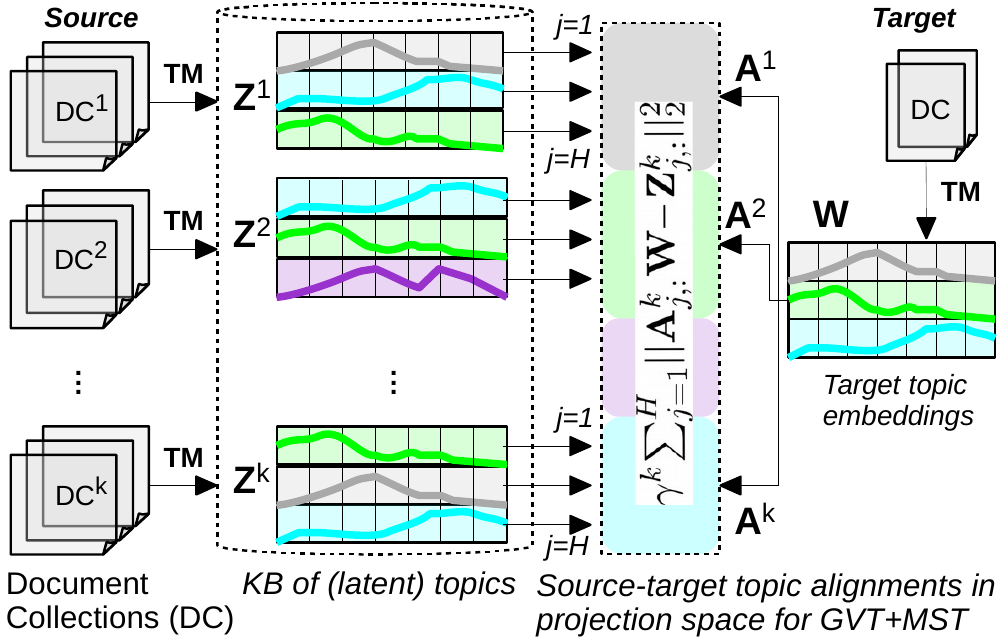}
  \caption{(Left) DocNADE (LVT+MST):  Introducing multi-source word embeddings at each autoregressive step $i$. % in computing $\log p({\bf v})$. 
% for a document ${\bf v}$ of size $D$. 
Double circle $\rightarrow$ multinomial (softmax) unit.
(Right) An illustration of (latent) topic alignments between source and target corpora in GVT+MST configuration. 
Each row in ${\bf Z}^{k}$ is a topic embedding that explains the underlying thematic structures of the source corpus, $DC^{k}$. 
Here, TM refers to DocNADE model. 
%LVT $\rightarrow$ Local View Transfer
}
  \label{fig:AutoregressiveNetworks}
\end{figure}

\iffalse
\begin{figure}[t]
  \centering
  \includegraphics[scale=.75]{./MSTGVTillustration.pdf}
  \caption{Illustration of (latent) topic alignments between source and target corpora in GVT+MST configuration. 
Each row in ${\bf Z}^{k}$ is a topic embedding that explains the underlying thematic structures of the source corpus, $DC^{k}$. 
Here, TM refers to DocNADE model.}
  \label{fig:MSTGVTillustration}
\end{figure}
\fi

\section{Knowledge Transfer in Neural Topic Modeling}
Consider a sparse target domain $\mathcal{T}$ and a set of $|\mathcal{S}|$ source domains $\mathcal{S}$, 
we first prepare two knowledge bases (KBs) of representations from each of the sources: 
(1) word embeddings matrices $\{ {\bf E}^1, ..., {\bf E}^{|\mathcal{S}|} \}$, where ${\bf E}^k \in \mathbb{R}^{E \times K}$ and 
(2) latent topic features $\{ {\bf Z}^1, ..., {\bf Z}^{|\mathcal{S}|} \}$, 
where ${\bf Z}^k \in \mathbb{R}^{H \times K}$ encodes a distribution over a vocabulary of $K$ words. 
$E$ and $H$ are word embedding and latent topic dimensions, respectively.  
%In a TM paradigm, we use the KBs to improve representations on $\mathcal{T}$. 
While topic modeling on $\mathcal{T}$, we introduce two types of knowledge transfers from one or several sources: 
{\it  Local} (\texttt{LVT}) and {\it Global} (\texttt{GVT}) {\it View Transfer} using the two KBs of (pre-trained) latent word and topic representations, respectively.   
We employ a neural autoregressive topic model (i.e., DocNADE \citep{DBLP:conf/nips/LarochelleL12}) to prepare the KBs.

{\it Notice} that a superscript indicates a source. See Table \ref{tab:notations} for the notations used in this work.

\subsection{Neural Autoregressive Topic Models} 
%{\bf Neural Autoregressive Topic Model}:  
DocNADE \citep{DBLP:conf/nips/LarochelleL12} is an unsupervised neural-network based topic model that is inspired by the 
benefits of NADE \citep{DBLP:journals/jmlr/LarochelleM11} and RSM \citep{DBLP:conf/nips/SalakhutdinovH09} architectures. 
RSM has difficulties due to intractability leading to approximate gradients of the negative log-likelihood, while NADE does not require such approximations.   
On other hand, RSM is a generative model of word count, while NADE is limited to binary data. 
Specifically, DocNADE factorizes the joint probability distribution of words in a document
as a product of conditional distributions and models each conditional via
a feed-forward neural network to efficiently compute a document representation. 
%following the NADE architecture. 

%Additionally, DocNADE has shown to outperform traditional models such as LDA \citep{DBLP:journals/jmlr/BleiNJ03} and RSM \citep{DBLP:conf/nips/SalakhutdinovH09} 
%in terms of both the log-probability on unseen documents and retrieval accuracy.
%therefore we adopt it to perform knowledge transfer within topic modeling framework.

%Since DocNADE \citep{DBLP:conf/nips/LarochelleL12}, a neural-network based topic model 
%has shown to outperform traditional models such as LDA \citep{DBLP:journals/jmlr/BleiNJ03} and RSM \citep{DBLP:conf/nips/SalakhutdinovH09} 
%in terms of both the log-probability on unseen documents and retrieval accuracy, 
%therefore we adopt it to perform knowledge transfer within topic modeling framework.
%to perform knowledge transfer.% within TM.
\begin{figure}[t]
 \centering
\begin{minipage}{.65\linewidth}
\begin{algorithm}[H]
\centering
\caption{{%\small 
Computation of $\log p({\bf v})$ and Loss $\mathcal{L}({\bf v})$}}\label{algo:computelogpv} %in \\ 
%{\it DocNADE}, {\it ctx-DocNADE} or {\it ctx-DocNADEe}}}
%{%\small 
\begin{algorithmic}
\Statex \textbf{Input}: A target training document ${\bf v}$, ${\mathcal{|S|}}$ source domains%,  Embedding matrix {\bf E} 
\Statex \textbf{Input}: KB of latent topics $\{{\bf Z}^{1}, ..., {\bf Z}^{\mathcal{|S|}}\}$
\Statex \textbf{Input}: KB of word embedding matrices $\{{\bf E}^{1}, ..., {\bf E}^{\mathcal{|S|}}\}$
\Statex \textbf{Parameters}: ${\bf \Theta} = \{{\bf b}, {\bf c}, {\bf W}, {\bf U},  {\bf A}^{1}, ...,  {\bf A}^{\mathcal{|S|}}\}$
\Statex \textbf{Hyper-parameters}: ${\bf \theta} = \{\lambda^{1}, ..., \lambda^{\mathcal{|S|}}, \gamma^{1}, ..., \gamma^{\mathcal{|S|}}, H$\}
%\Statex \textbf{Output}: $\log p({\bf v})$
%\State \textbf{Parameters}: {\it RSM parameters}: \{$W_{vh}$, $b_{v}$, $b_{v}^{(t)}$, $b_{h}$, $b_{h}^{(t)}$\}
%\State \textbf{Parameters}: {\it RNN Connections}: \{$W_{uh}, $ $W_{uv},  W_{uu},  W_{vu}, u^{0}$\}
%\Statex \textbf{RNN-RSM Parameters}: 
\State Initialize: ${\bf a} \gets {\bf c}$ and $ p({\bf v}) \gets 1$
%\State $ p({\bf v}) = 1$
\For{$i$ from $1$ to $D$}
        %\State compute ${\bf h}_{i} ({\bf v}_{<i})  \gets g({\bf a}$), where $g$ = \{sigmoid, tanh\}
        \State  ${\bf h}_{i} ({\bf v}_{<i})  \gets g({\bf a}$), where $g$ = \{sigmoid, tanh\}
        \State $p(v_{i}=w | {\bf v}_{<i}) \gets \frac{\exp (b_w + {\bf U}_{w,:} {\bf h}_i ({\bf v}_{<i}))}{\sum_{w'} \exp (b_{w'} + {\bf U}_{w',:} {\bf h}_i ({\bf v}_{<i}))}$
	\State $ p({\bf v}) \gets  p({\bf v}) p(v_{i} | {\bf v}_{<i})$
        \State compute pre-activation at step, $i$: ${\bf a} \gets {\bf a} + {\bf W}_{:, v_{i}}$  
        \If  {\texttt{LVT}} 
             \State get word embedding for $v_i$ from source domain(s) 
             \State ${\bf a} \gets {\bf a} + \sum_{k=1}^{\mathcal{|S|}} \lambda^k \ {\bf E}_{:, v_{i}}^{k}$
        \EndIf
\EndFor
\State $\mathcal{L}({\bf v}) \gets  - \log p({\bf v})$

\If  {\texttt{GVT}} 
         %\State $\mathcal{L} \gets \mathcal{L} + \sum_{k=1}^{|s|} \gamma_k  \ || {\bf A} {\bf W} - {\bf Z}^{k} ||_2$
         %\State loss, $\mathcal{L} \gets \mathcal{L} + \sum_{k=1}^{|s|} \gamma_k  \ || {\bf A}^{k} {\bf W} - {\bf Z}^{k} ||_2$
         \State $\mathcal{L}({\bf v}) \gets \mathcal{L}({\bf v}) + \sum_{k=1}^{\mathcal{|S|}} \gamma^k  \ \sum_{j=1}^{H}  ||  {\bf A}_{j,:}^{k} {\bf W} - {\bf Z}_{j, :}^{k} ||_2^2$
\EndIf
\end{algorithmic}%}
\end{algorithm}
\end{minipage}
\end{figure}

{\bf DocNADE Formulation}: 
For a document ${\bf v}$ = $(v_1, ..., v_D)$ of size $D$, each word index $v_i$ takes value in $\{1, ..., K\}$ of vocabulary size $K$.  
DocNADE learns topics %over a sequence of words $(v_1, ...v_D)$ in a document ${\bf v}$
 in a language modeling fashion \citep{DBLP:journals/jmlr/BengioDVJ03} and 
decomposes the joint distribution $p({\bf v})$=$\prod_{i=1}^{D} p(v_i | {\bf v}_{<i})$ such that 
each autoregressive conditional $p(v_i | {\bf v}_{<i})$ is modeled by a feed-forward neural network 
using preceding words ${\bf v}_{<i}$ in the sequence:
\begin{equation*}\label{eq:DocNADEconditionals}
%\begin{split}
{\bf h}_i({\bf v}_{<i})  =  g ({\bf c} + \sum_{q<i} {\bf W}_{:, v_q})  \ \ \mbox{     and     } \ \ 
p (v_i = w | {\bf v}_{<i})  = \frac{\exp (b_w + {\bf U}_{w,:} {\bf h}_i ({\bf v}_{<i}))}{\sum_{w'} \exp (b_{w'} + {\bf U}_{w',:} {\bf h}_i ({\bf v}_{<i}))}
%\end{split}
\end{equation*}
for $i \in \{1,...D\}$, where ${\bf v}_{<i}$ is the subvector consisting of all $v_q$ such that $q < i$ i.e., ${\bf v}_{<i} \in \{v_1, ...,v_{i-1}\}$, 
$g(\cdot)$ is a non-linear activation function, ${\bf W} \in \mathbb{R}^{H \times K}$ and ${\bf U} \in \mathbb{R}^{K \times H}$ 
are weight %parameter 
matrices, ${\bf c} \in \mathbb{R}^H$ and ${\bf b} \in \mathbb{R}^K$ are bias parameter 
vectors. $H$ is the number of hidden units (topics). 
 %See an illustration in Figure \ref{fig:AutoregressiveNetworks}. 
%Figure \ref{fig:AutoregressiveNetworks} provides an illustration of the DocNADE model. 

%For DocNADE, Figure \ref{fig:AutoregressiveNetworks} and Algorithm \ref{algo:computelogpv} (\texttt{LVT} and \texttt{GVT} set to {\it False})
%demonstrate the computation of $\log p({\bf v})$ and 
%negative log-likelihood $\mathcal{L}({\bf v})$ %  =  - \sum_{i=1}^{D} \log p (v_i | {\bf v}_{<i})$   of ${\bf v}$. 
%that is 
%minimized using %(stochastic) 
%gradient descent. % to learn {$\bf \Theta$}.

%{\bf Why DocNADE as a TM framework for knowledge transfers via word and topic features}:   
Figure \ref{fig:AutoregressiveNetworks} (left) (without ``KB of word embeddings") provides an illustration of the $i$th autoregressive step of the DocNADE architecture, where 
the parameter $\bf W$ is shared in the feed-forward networks and ${\bf h}_i$ encodes topic-proportion embedding.  Importantly, the topic-word matrix $\bf W$ has a property that  
the column vector ${\bf W}_{:, v_i}$ corresponds to embedding of the word $v_i$, 
whereas the row vector ${\bf W}_{j, :}$ encodes latent features for $j$th topic. 
We leverage this property to introduce external knowledge via latent word and topic features. 

Additionally, DocNADE has shown to outperform traditional models such as LDA \citep{DBLP:journals/jmlr/BleiNJ03} and RSM \citep{DBLP:conf/nips/SalakhutdinovH09} 
in terms of both the log-probability on unseen documents and retrieval accuracy.
Recently, \citet{pankajgupta:2019iDocNADEe} has improved topic modeling on short texts by introducing word embeddings \citep{D14-1162} in DocNADE architecture. 
Thus, we adopt DocNADE to perform knowledge transfer within the neural topic modeling framework.
  
%Importantly, we exploit properties of $\bf W$ in DocNADE that 
%the column vector ${\bf W}_{:, v_i}$ corresponds to embedding of the word $v_i$, 
%whereas the row vector ${\bf W}_{j, :}$ encodes latent features for $j$th topic. 
%Therefore, we apply DocNADE to prepare KBs of ${\bf E}$ and ${\bf Z}$ using source domains $\mathcal{S}$.  

%Additionally, DocNADE has shown to outperform traditional models such as LDA \citep{DBLP:journals/jmlr/BleiNJ03} and RSM \citep{DBLP:conf/nips/SalakhutdinovH09} 
%in terms of both the log-probability on unseen documents and retrieval accuracy.
%therefore we adopt it to perform knowledge transfer within topic modeling framework.
% aslo docnadee.....

Algorithm \ref{algo:computelogpv} (for DocNADE, set \texttt{LVT} and \texttt{GVT} to {\it False})
demonstrates the computation of $\log p({\bf v})$ and 
negative log-likelihood $\mathcal{L}({\bf v})$ %  =  - \sum_{i=1}^{D} \log p (v_i | {\bf v}_{<i})$   of ${\bf v}$. 
that is minimized using %stochastic 
gradient descent. % to learn {$\bf \Theta$}.
%The linear complexity
Moreover, computing ${\bf h}_i$ is efficient (linear complexity)
due to the NADE architecture that leverages the pre-activation ${\bf a}_{i-1}$ of $(i-1)$th
step in computing the pre-activation ${\bf a}_i$ for the $i$th step. 
See \citet{DBLP:conf/nips/LarochelleL12} for further details.

%\subsubsection*{Fine-granularity (Local View) Transfers}
\subsection{Multi-View (MVT) and Multi-Source Transfers (MST) in Topic Modeling}
%\subsection{MVT and MST in Topic Modeling}
% split the para into local, global ang multi-view an then mention MST
Here, we describe a topic modeling framework that jointly exploits the complementary knowledge 
using the two KBs of (pre-trained) latent word and topic representations (or embeddings), 
obtained from large document collections (DCs) from several sources.  
In doing so, we first apply the DocNADE to generate a topic-word matrix for each of the DCs, 
where its column-vector and row-vector generate ${\bf E}^{k}$ and ${\bf Z}^{k}$, respectively for the $k$th source. 

{\bf LVT+MST Formulation}:  
As illustrated in Figure \ref{fig:AutoregressiveNetworks} (left) and Algorithm \ref{algo:computelogpv} with \texttt{LVT}={\it True}, 
we perform knowledge transfer to a target $\mathcal{T}$ using a KB of pre-trained word embeddings $\{ {\bf E}^1, ..., {\bf E}^{|\mathcal{S}|} \}$ 
from several sources $\mathcal{S}$ (i.e., multi-source): 
\begin{equation*}\label{eq:LVTMST}
{\bf h}_i({\bf v}_{<i})  =  g ({\bf c} + \sum_{q<i} {\bf W}_{:, v_q} +  \sum_{q<i} \sum_{k=1}^{\mathcal{|S|}} \lambda^k \ {\bf E}_{:, v_{q}}^{k} ) 
\end{equation*}
Here, $k$ refers to the $k$th source and 
$\lambda^{k}$ is a weight for %the $k$th source embeddings i.e., 
${\bf E}^{k}$ that %that determines its relevance for $\mathcal{T}$ 
%and therefore, controls the amount of knowledge transferred.  
controls the amount of knowledge transferred in $\mathcal{T}$, based on domain overlap between target and source(s).  
Recently, DocNADEe \citep{pankajgupta:2019iDocNADEe} has incorporated word embeddings \citep{D14-1162} in extending 
DocNADE; however, it is based on a single source. 

{\bf GVT+MST Formulation}:  
Next, we perform knowledge transfer exclusively using the KB of pre-trained latent topic features (e.g., ${\bf Z}^k$) from one or several sources, $\mathcal S$. 
%Figure \ref{fig:MSTGVTillustration} 
%as described in Algorithm \ref{algo:computelogpv} (when \texttt{GVT} = {\it True}).  
%(Algorithm \ref{algo:computelogpv} when \texttt{GVT} = {\it True}).
In doing so, we add a regularization term to the loss function $\mathcal{L}({\bf v})$ and require 
%we minimize it in such a way that 
DocNADE to minimize the overall loss in a way that the (latent) topic features in  ${\bf W}$
simultaneously inherit relevant topical features from each of the source domains $\mathcal{S}$, 
and generate meaningful representations for the target $\mathcal{T}$. % in order to address data-sparsity. 
%and minimizes  the negative likelihood of ${\bf v}$ in $\mathcal{T}$.  
The overall loss $\mathcal{L}({\bf v}) $ due to GVT+MST in DocNADE is given by: 
\begin{equation*}
\mathcal{L}({\bf v}) =  - \log p({\bf v}) + \sum_{k=1}^{\mathcal{|S|}} \gamma^k  \ \sum_{j=1}^{H}  ||  {\bf A}_{j,:}^{k} {\bf W} - {\bf Z}_{j, :}^{k} ||_2^2
\end{equation*}
Here, ${\bf A}^{k}$$\in$$\mathbb{R}^{H \times H}$ aligns latent topics in  the target $\mathcal{T}$ and $k$th source,  
and $\gamma^{k}$ governs the degree of imitation of topic features ${\bf Z}^{k}$ %of $k$th source 
by ${\bf W}$ in $\mathcal{T}$. 
%signifies relevance of $k$th source. 
Consequently, the generative process of learning meaningful (latent) topic features in ${\bf W}$ 
is guided by relevant features in $\{{\bf Z}\}_{1}^{\mathcal{|S|}}$ to address data-sparsity. % in order to get meaningful representations for $\mathcal{T}$.   
Algorithm \ref{algo:computelogpv} describes the computation of the loss, when \texttt{GVT} = {\it True} and \texttt{LVT} = {\it False}. 

Moreover, Figure \ref{fig:AutoregressiveNetworks} (right) illustrates the need for topic alignments between target and source(s). Here, $j$ indicates the topic (i.e., row) index in a topic matrix, e.g., ${\bf Z}^{k}$.   
Observe that the first topic (gray curve), i.e., $Z_{j=1}^{1} \in {\bf Z}^{1}$ of the first source aligns with the first row-vector (i.e., topic) of ${\bf W}$ (of target).  
However, the other two topics  $Z_{j=2}^{1},  Z_{j=3}^{1} \in {\bf Z}^{1}$ need alignment with the target topics.  

%Notice that 
%Figure \ref{fig:MSTGVTillustration} 
%as described in Algorithm \ref{algo:computelogpv} (when \texttt{GVT} = {\it True}).  
%(Algorithm \ref{algo:computelogpv} when \texttt{GVT} = {\it True}).

{\bf MVT+MST Formulation}:  
When \texttt{LVT} and \texttt{GVT} are {\it True} (Algorithm \ref{algo:computelogpv}) for many sources, 
the two complementary representations are jointly used in knowledge 
transfer and therefore, the name {\it multi-view} and {\it multi-source} transfers.

\begin{table*}[t]
    %\caption{Global caption}
    \begin{minipage}{.76\linewidth}
      \center
	\renewcommand*{\arraystretch}{1.25}
	\resizebox{.995\linewidth}{!}{
         \setlength\tabcolsep{2.5pt}
       \begin{tabular}{r|r|rrrrrr||r|r|rrrrrr}
%\hline
\multicolumn{8}{c||}{\texttt{Target Domain Corpora}} & \multicolumn{8}{c}{\texttt{Source Domain Corpora}} \\ %\hline
 \multicolumn{1}{c|}{\bf ID} & \multicolumn{1}{c|}{\bf Data} &  \multicolumn{1}{c}{\bf Train} &  \multicolumn{1}{c}{\bf Val} & \multicolumn{1}{c}{\bf Test} &  \multicolumn{1}{c}{$K$}   & \multicolumn{1}{c}{\bf L} & \multicolumn{1}{c||}{\bf C}     
&   \multicolumn{1}{c|}{\bf ID} &  \multicolumn{1}{c|}{\bf Data} &  \multicolumn{1}{c}{\bf Train} &  \multicolumn{1}{c}{\bf Val} & \multicolumn{1}{c}{\bf Test} &  \multicolumn{1}{c}{$K$}   & \multicolumn{1}{c}{\bf L} & \multicolumn{1}{c}{\bf C}    \\ \hline 
$\mathcal{T}^1$	&    20NSshort             & 1.3k       & 0.1k        & 0.5k           &    1.4k           &  13.5        &   20                   	&  $\mathcal{S}^1$  	& 20NS                  & 7.9k & 1.6k & 5.2k  &     2k          &   107.5     &   20        \\
$\mathcal{T}^2$	&    20NSsmall             & 0.4k      &  0.2k       & 0.2k            &      2k             &   187.5     &    20                 	&  $\mathcal{S}^2$  	&  R21578              &  7.3k&  0.5k & 3.0k  &      2k               &   128       &   90       \\
$\mathcal{T}^3$	&    TMNtitle                & 22.8k    &  2.0k     & 7.8k              &       2k           &   4.9          &     7                  	&  $\mathcal{S}^3$  	&   TMN                 & 22.8k  &  2.0k & 7.8k  &      2k         &    19          &       7     \\
$\mathcal{T}^4$	&    R21578title            &  7.3k     &  0.5k     & 3.0k            &   2k                  &   7.3          &     90               	&  $\mathcal{S}^4$  	& AGNews              & 118k & 2.0k &   7.6k &        5k         &  38       &     4         \\
$\mathcal{T}^5$	&    Ohsumedtitle         &  8.3k     &  2.1k     & 12.7k            &   2k                  &   11.9          &     23               &  $\mathcal{S}^5$  & PubMed             &  15.0k     &  2.5k     & 2.5k            &   3k                  &   254.8          &     -    \\
$\mathcal{T}^6$  	& Ohsumed          &  8.3k     &  2.1k     & 12.7k            &   3k                  &   159.1          &     23 			&              &       &      &           &                     &            &        \\  	             
\end{tabular}}
\caption{Data statistics: Short/long texts and/or small/large corpora in target and source domains.  
Symbols-  $K$: vocabulary size,  $L$: average text length (\#words), $C$: number of classes and  $k$: thousand. 
For short-text,  $L$$<$$15$. $\mathcal{S}^3$ is also used in target domain.}
\label{tab:datadescription}
    \end{minipage} \ \ %
\begin{minipage}{.22\linewidth}
      \centering
	\renewcommand*{\arraystretch}{1.2}
	\resizebox{.995\linewidth}{!}{
	\setlength\tabcolsep{2.5pt}
        \begin{tabular}{c||c|c|c|c|c|c} %\hline
				& $\mathcal{T}^1$	& $\mathcal{T}^2$ 	& $\mathcal{T}^3$ 	& $\mathcal{T}^4$ 	& $\mathcal{T}^5$ 	 & $\mathcal{T}^6$  \\ \hline \hline
$\mathcal{S}^1$	& $\mathcal{I}$ 			& $\mathcal{I}$  			& $\mathcal{R}$ 			& $\mathcal{D}$  			&  $\mathcal{D}$ 	  		&  $\mathcal{D}$	  \\ \hline
$\mathcal{S}^2$	& $\mathcal{D}$  			& $\mathcal{D}$ 			& $\mathcal{D}$  			& $\mathcal{I}$  			&  $\mathcal{D}$ 	  		&  $\mathcal{D}$	  \\ \hline
$\mathcal{S}^3$	& $\mathcal{R}$  			& $\mathcal{R}$ 			& $\mathcal{I}$				& $\mathcal{D}$ 			&  $\mathcal{D}$ 	  		&  $\mathcal{D}$	  \\ \hline
$\mathcal{S}^4$	& $\mathcal{R}$ 			& $\mathcal{R}$ 			& $\mathcal{R}$ 			& $\mathcal{D}$			&  $\mathcal{D}$			&  $\mathcal{D}$ 	  \\ \hline
$\mathcal{S}^5$	& $\mathcal{D}$  			& $\mathcal{D}$ 			& $\mathcal{D}$ 			& $\mathcal{D}$			& {-}		  		& {-} 	  %\\ \hline
\end{tabular}}
\caption{Domain overlap in source-target corpora. $\mathcal{I}$: Identical,  $\mathcal{R}$: Related and $\mathcal{D}$: Distant domains.}
\label{tab:domainoverlap}
    \end{minipage} 
\end{table*}

\iffalse
    \begin{minipage}{.22\linewidth}
      \centering
	\renewcommand*{\arraystretch}{1.2}
	\resizebox{.995\linewidth}{!}{
	\setlength\tabcolsep{2.5pt}
        \begin{tabular}{c||c|c|c|c|c|c} %\hline
				& $\mathcal{S}^1$	& $\mathcal{S}^2$ 	& $\mathcal{S}^3$ 	& $\mathcal{S}^4$ 	& $\mathcal{S}^5$ 	 & $\mathcal{S}^6$  \\ \hline \hline
$\mathcal{T}^1$	& $\mathcal{I}$ 			& $\mathcal{D}$  			& $\mathcal{R}$ 			& $\mathcal{R}$  			&  $\mathcal{D}$ 	  		&  $\mathcal{D}$	  \\ \hline
$\mathcal{T}^2$	& $\mathcal{I}$  			& $\mathcal{D}$ 			& $\mathcal{R}$  			& $\mathcal{R}$  			&  $\mathcal{D}$ 	  		&  $\mathcal{D}$	  \\ \hline
$\mathcal{T}^3$	& $\mathcal{R}$  			& $\mathcal{D}$ 			& $\mathcal{I}$				& $\mathcal{R}$ 			&  $\mathcal{D}$ 	  		&  $\mathcal{D}$	  \\ \hline
$\mathcal{T}^4$	& $\mathcal{D}$ 			& $\mathcal{I}$ 			& $\mathcal{D}$ 			& $\mathcal{D}$			&  $\mathcal{D}$			&  $\mathcal{D}$ 	  \\ \hline
$\mathcal{T}^5$	& $\mathcal{D}$  			& $\mathcal{D}$ 			& $\mathcal{D}$ 			& $\mathcal{D}$			& $\mathcal{I}$		  		& {-} 	  %\\ \hline
\end{tabular}}
\caption{Domain overlap in target-source corpora. $\mathcal{I}$: Identical,  $\mathcal{R}$: Related and $\mathcal{D}$: Distant domains.}
\label{tab:domainoverlap}
\fi

\section{Evaluation and Analysis} 
{\bf Datasets}: Table \ref{tab:datadescription} describes %provides details of 
the datasets used in high-resource source  
and low-and high-resource target domains for our experiments. 
The target domain ${\mathcal T}$ consists of four short-text corpora ({\small \texttt{20NSshort}}, {\small \texttt{TMNtitle}}, {\small \texttt{R21578title}} and {\small \texttt{Ohsumedtitle}}),
one small corpus ({\small \texttt{20NSsmall}}) and two large corpora ({\small \texttt{TMN}} and {\small \texttt{Ohsumed}}).  
However in source $\mathcal{S}$, we use five large corpora ({\small \texttt{20NS}}, {\small \texttt{R21578}}, {\small \texttt{TMN}}, {\small \texttt{AGnews}} and {\small \texttt{PubMed}}) 
in different label spaces (i.e, domains). Here, the corpora ($\mathcal{T}^5$, $\mathcal{T}^6$ and $\mathcal{S}^5$) belong to {\it medical} and others to {\it news}.

Additionally, Table \ref{tab:domainoverlap} suggests domain overlap (in terms of label match) in the target and source corpora, 
where we define three types of overlap: $\mathcal{I}$ (identical) if all labels match, $\mathcal{R}$ (related) if some labels match, and $\mathcal{D}$ (distant) if a very few or no labels match.
Note, our modeling approaches are completely unsupervised and do not use the data labels. 
See the data labels in {\it supplementary}. 

{\bf Baselines}: As summarized in Table \ref{tab:baselinesvsthiswork}, we consider several baselines including 
(1) LDA-based and neural network-based topic models that use the target data, 
(2) topic models using pre-trained word embeddings (i.e., LVT) from \citet{D14-1162} (Glove), 
(3) unsupervised document representation, where we employ doc2vec \citep{DBLP:conf/icml/LeM14} and EmbSum 
(to represent a document by summing the embedding vectors of it’s words using Glove) 
in order to quantify the quality of document representations, 
(4) zero-shot topic modeling, where we use all source corpora and no target corpus, and 
(5) data-augmentation, where we use all source corpora along with a target corpus for TM on $\mathcal{T}$. 
%We consider topic models, e.g.,  
%(1) glove-DMM \citep{DBLP:journals/tacl/NguyenBDJ15}: LDA-based with word embedding    
%(2) DocNADE:  Neural network-based, and (3) DocNADEe \citep{pankajgupta:2019iDocNADEe}: DocNADE+Glove embeddings \citep{D14-1162}. 
Using DocNADE, we first prepare two KBs of word embeddings and latent topics from each of the source corpora, 
and then use them in knowledge transfer to ${\mathcal T}$. 
%See the {\it supplementary} for experimental setup and Hyperparameters. %ablation study.
%See the configuration of different hyper-parameters in supplementary. 

{\bf Reproducibility}:
% hyperparameters 
% code released 
% data preprocessing scripts and preprocessed data released 
For evaluations in the following sections, we follow the experimental setup similar to DocNADE \citep{DBLP:conf/nips/LarochelleL12} and DocNADEe \citep{pankajgupta:2019iDocNADEe}, 
where the number of topics ($H$) is set to $200$. 
See {\it supplementary} for the experimental setup, hyperparameters\footnote{selected with grid search;  suboptimal results (see {\it supplementary}) by learning $\lambda$ and $\gamma$ with backpropagation} 
and optimal values of $\lambda^k \in [0.1, 0.5, 1.0]$ and $\gamma^k \in [0.1, 0.01, 0.001]$ (determined using development set) %in corresponding source and target configurations.  
in different source-target configurations.  
%See the experimental setup and hyper-parameter configurations in {\it supplementary}. 
In addition, we provide the code. 

\begin{table*}[t]
\center
\renewcommand*{\arraystretch}{1.3}
\resizebox{.75\textwidth}{!}{
\begin{tabular}{r|c|c|c|c}

\multicolumn{1}{c|}{\multirow{2}{*}{\bf Baselines (Related Works)}}    &    \multicolumn{4}{c}{\bf Features} \\ \cline{2-5}
  										& \textit{NTM}		 & \textit{AuR}  & \textit{LVT} & \textit{GVT}$|$\textit{MVT}$|$\textit{MST} \\ \hline
LDA \citep{DBLP:journals/jmlr/BleiNJ03} 	      		&		 		&			&				  &		    			 \\ 
RSM 	\citep{DBLP:conf/nips/SalakhutdinovH09}	        	&	\checkmark	&			&		  &		     			\\ 
DocNADE 	\citep{DBLP:conf/nips/LarochelleL12}	        &	\checkmark	&	\checkmark&		  &		     			\\ 
NVDM \citep{DBLP:conf/icml/MiaoYB16}	      			&	\checkmark	&			 &		  &		     			\\ 
ProdLDA \citep{SrivastavaSutton}    	      			 &		 		&			&		  &		    			\\ \hdashline

Gauss-LDA \citep{P15-1077} 	      				 &		 		&			&	\checkmark	  &		    			 \\ 		
glove-DMM \citep{DBLP:journals/tacl/NguyenBDJ15}  	 &		 		&			&	\checkmark	  &		    			\\ 
DocNADEe  \citep{pankajgupta:2019iDocNADEe}     	&	\checkmark	&	\checkmark&	\checkmark	  &		     			\\  \hdashline

EmbSum	      								&		 		&			&		  &		     			\\ 
doc2vec 	\citep{DBLP:conf/icml/LeM14}   	    		&		 		&			&		  &		     			\\  \hdashline

{\bf this work}	   					&	\checkmark	 		&	\checkmark&	\checkmark	  &	\checkmark  \quad  \checkmark  \quad   \checkmark	%\\   \hline 
 
\end{tabular}}
\caption{Baselines (related works) vs this work. Here, {\it NTM} and {\it AuR} refer to neural network-based TM and autoregressive assumption, respectively.  
DocNADEe $\rightarrow$  DocNADE+Glove embeddings. 
%Note that {EmbSum} and {doc2vec} are document representation baselines and are not topic models.
}
\label{tab:baselinesvsthiswork}
\end{table*}

\begin{table*}[t]
\center
%\small
\renewcommand*{\arraystretch}{1.3}
\resizebox{.999\textwidth}{!}{
\setlength\tabcolsep{3.5pt}
\begin{tabular}{r|r|ccc|ccc|ccc|ccc||cc}

\multirow{1}{*}{\bf KBs from}   &     \multirow{1}{*}{\bf Model/} & \multicolumn{14}{c}{{\bf Scores on Target Corpus} ({\it in sparse-data  and sufficient-data settings})}   \\ \cline{3-16}

\multirow{1}{*}{\bf Source}      &   \multirow{1}{*}{\bf Transfer} &   \multicolumn{3}{c|}{\texttt{20NSshort}}     &  \multicolumn{3}{c|}{\texttt{TMNtitle}}   &   \multicolumn{3}{c|}{\texttt{R21578title}}   &    \multicolumn{3}{c||}{\texttt{20NSsmall}}  &    \multicolumn{2}{c}{\texttt{TMN}}\\
   \multirow{1}{*}{\bf Corpus}   &   \multirow{1}{*}{\bf Type}  &   $PPL$  &  $COH$ &  $IR$ &    $PPL$  &  $COH$ &  $IR$ &  $PPL$  &  $COH$ &  $IR$ & $PPL$  &  $COH$ &  $IR$   & $PPL$  &  $COH$ \\ \hline % &  $IR$ \\ \hline
% &   &   PPL  &  COH &  IR &    PPL  &  COH &  IR &  PPL  &  COH &  IR & PPL  &  COH &  IR   \\ \hline

\multirow{1}{*}{\it Baseline TM}   &  NVDM    &  1047   & {\bf .736}   & .076     &  973   & .740   & .190     &  372   & .735   & .271     &  957   & .515   & .090    &  833   & .673  \\ % & .000 \\
\multirow{1}{*}{\it {\bf without} Word-} &  ProdLDA    &  923   & .689   & .062     &  1527   & .744   & .170     &  480   & {\bf .742}   & .200     &  1181   & .394   & .062     &  1519   & .577  \\   % & .000 \\ % say in footnote, DocNADE outperfroms LDA and RSM. ProdLDA outperforms NVDM.
\multirow{1}{*}{\it Embeddings}	    &   DocNADE    &  646   & .667   & .290     &  706   & .709   & .521     &  192   & .713   & .657     &  594   & .462   & .270   &  584   & .636 \\ \hline \hline  %& .000  \\  \hline \hline % & .000 \\  \hline \hline

\multirow{3}{*}{\it 20NS}    &  LVT     &  630   & .673   & .298     &  705   & .709   & .523     &  194   & .708   & .656     &  594   & .455   & .288     		&  582   & .649 \\%  & .653\\ %.757
					  &  GVT     &  646   & .690   & .303     &  718   & .720   & .527     &  {\bf 184}   & .698   & .660     &  594   & .500   & .310     &  590   & .652  \\% & .653\\  %.752
					  &  MVT &  {\bf 638}   & .690   & {\bf .314}     &  714   & .718   & .528     &  188   & .715   & .655     &  600   & .499   & .311    	 &  588   & .650   \\ \hline %& .653\\ %.760

\multirow{3}{*}{\it TMN}    &  LVT     &  649   & .668   & .296     &  {\bf 655}   & .731   & .548     &  187   & .703   & .659     &  593   & .460   & .273    	&  -   & -   \\ %& - \\
					  &  GVT          &  661   & .692   & .294     &  689   & .728   & .555     &  191   & .709   & .660     &  596   & .521   & .276     	&  -   & -   \\ %& -\\
					  &  MVT   &  658   & .687   & .297     &  663   & .747   & .553     &  195   & .720   & .660     &  599   & .507   & .292     		&  -   & -   \\ \hline %& -\\ 
					 
\multirow{3}{*}{\it R21578}  &  LVT      &  656   & .667   & .292     &  704   & .715   & .522     &  186   & .715   & .676     &  593   & .458   & .267     	&  581   & .636  \\ % & .652\\ %.759
					   &  GVT          &  654   & .672   & .293     &  716   & .719   & .526     &  194   & .706   & .672     &  595   & .485   & .279   	&  591   & .646   \\ %& .653  \\   %.757
					  &  MVT    &  650   & .670   & .296     &  716   & .720   & .528     &  194   & .724   & {.676}     &  599   & .490   & .280     	&  589   & .650  \\ \hline % & .653\\  %.759
					 
\multirow{3}{*}{\it AGnews}  &  LVT      &  650   & .677   & .297     &  682   & .723   & .533     &  185   & .710   & .659     &  {\bf 592}   & .458   & .260     	&  {\bf 564}   & .668  \\ % & .652\\  %.761
					   &  GVT          &  667   & .695   & .300     &  728   & .735   & .534     &  190   & .717   & .663     &  598   & .563   & .282   	&  601   & .684   \\ %& .653  \\  %.767
					  &  MVT    &  659   & .696   & .290     &  718   & .740   & .533     &  189   & .727   & .659     &  599   & .566   & .279     		&  592   & {.686}  \\  \hline \hline % & .653\\  %.767
					 
\multirow{3}{*}{\it MST}  &  LVT      &  640   & .678   & .308     &  663   & .732   & .547     &  186   & .712   & .673     &  596   & .442   & .277   								&  568   	 & .674 \\%  & .653  \\  %.757
					   &  GVT          &  658   & .705   & .305     &  704   & .746   & .550     &  192   & .727   & .673     &  599   & .585   & {\bf .326} 						&  602   & .680  \\ % & .654\\   %.762
					  &  MVT    &  656   & {.721}   & {\bf .314}     &  680   & {\bf .752}   & {\bf .556}     &  188   & {.738}   & {\bf .678}       &  600   & {\bf .600}   & .285    &  600   & {\bf .690}   \\ \hline \hline% & .653   \\   %.760
					  
\multicolumn{2}{r|}{{\bf Gain\%} (vs DocNADE)}    &  1.23   & {8.10}   & {8.28}     &  7.22   & {6.06}   & {6.72}     &  4.17   & {3.51}   & 3.12       &  0.34   & {29.87}   & 20.74    &  3.42   & {8.50}

\end{tabular}}
\caption{State-of-the-art comparisons with topic models: Perplexity (PPL), topic coherence (COH) and precision (IR) at retrieval fraction $0.02$. 
%KB: Knowledge-base of word and latent topic features. 
Scores are reported on each of the target, given KBs from one or several sources.  %given the word embedding and/or latent topic transfer(s) from one or more related or distant source domains.  
%LVT, GVT and MVT: {\it L}ocal, {\it G}lobal and {\it M}ulti View Transfer; 
%MST: Multi Source Transfer and 
%+ Glove: MVT+Glove embeddings. 
{\it Please read column-wise}. {\bf Bold}: best in column. %\underline{underline}: Scores higher than DocNADEe. 
 %View Transfer, GVT: Topic Transfer.  +Glove: EmbTF+TTF including Glove embeddings. ALL: multi-source transfer. TO DO: Replace EmbTF+TTF with MVT
}
\label{tab:scoresTMwithoutWordEmbeddings}
\end{table*}

\begin{table*}[t]
\center
%\small
\renewcommand*{\arraystretch}{1.3}
\resizebox{.999\textwidth}{!}{
\setlength\tabcolsep{3.5pt}
\begin{tabular}{r|r|ccc|ccc|ccc|ccc||cc}

\multirow{1}{*}{\bf KBs from}   &     \multirow{1}{*}{\bf Model/} & \multicolumn{14}{c}{{\bf Scores on Target Corpus} ({\it in sparse-data  and sufficient-data settings})}   \\ \cline{3-16}

\multirow{1}{*}{\bf Source}      &   \multirow{1}{*}{\bf Transfer} &   \multicolumn{3}{c|}{\texttt{20NSshort}}     &  \multicolumn{3}{c|}{\texttt{TMNtitle}}   &   \multicolumn{3}{c|}{\texttt{R21578title}}   &    \multicolumn{3}{c||}{\texttt{20NSsmall}}  &    \multicolumn{2}{c}{\texttt{TMN}}\\
   \multirow{1}{*}{\bf Corpus}   &   \multirow{1}{*}{\bf Type}  &   $PPL$  &  $COH$ &  $IR$ &    $PPL$  &  $COH$ &  $IR$ &  $PPL$  &  $COH$ &  $IR$ & $PPL$  &  $COH$ &  $IR$   & $PPL$  &  $COH$ \\ \hline % &  $IR$ \\ \hline
% &   &   PPL  &  COH &  IR &    PPL  &  COH &  IR &  PPL  &  COH &  IR & PPL  &  COH &  IR   \\ \hline

						 	&  doc2vec        &  -   & -       & .090     &  -   & -        & .190      &  -   & -        & .518     &  -   & -   & .200    &  -   & -   \\ % & .000\\  
      							&  EmbSum    &  -   & -   & .236     &  -   &  -   & .513     &  -   &  -   & .587     &  -   & -   & .214     &  -   & -   \\  \cdashline{1-16} % & .000\\  \cdashline{2-16}
\multirow{1}{*}{\it Baseline TM}    &  Gauss-LDA    &  -   & -   & .080     &  -   & -   & .408     &  -   & -   & .367     &  -   & -   & .090     &  -   & - \\ %  & .000\\ 
\multirow{1}{*}{\it {\bf with} Word-} 	      &  glove-DMM    &  -   & .512   & .183     &  -   & .633   & .445     &  -   & .364   & .273     &  -   & .578   & .090   &  -   & .705   \\ %& .000 \\                                    
\multirow{1}{*}{\it Embeddings}		&  DocNADEe    &  {\bf 629}   & .674   & .294     &  680   & .719   & .540     &  187   & .721   & .663     &  {\bf 590}   & .455   & .274   &  {\bf 572}  & .664 \\ \hline %& .000  \\  \hline \hline % & .000 \\  \hline \hline

\multirow{1}{*}{\it 20NS}    &  MVT+Glove    &  630   & .700   & {\bf .300}     &  690   & .733   & .539     &  {\bf 186}   & .724   & .664     &  601   & .499   & .306    	&  580   & .667  \\  % & .645\\  \hline %.773  \hline

\multirow{1}{*}{\it TMN}   &  MVT+Glove     &  640   & .689   & .295     &  {\bf 673}   & .750   & {\bf .543}     &  {\bf 186}   & .716   & .662     &  599   & .517   & .261    	 &  -   & -  \\  % & -\\  \hline

\multirow{1}{*}{\it R21578}     &  MVT+Glove     &  633   & .691   & .295     &  689   & .734   & .540     &  188   & {\bf .734}   & {\bf .676}     &  598   & .485   & .255    	&  580   & .670  \\  % & .645 \\  \hline %.773 \hline

\multirow{1}{*}{\it AGnews}   &  MVT+Glove    &  642   & .707   & .291     &  706   & .745   & .540     &  190   & {\bf .734}   & .664     &  600   & .573   & .284     	&  600   & {.690}   \\  \hline % & .646\\   \hline \hline  %.773

\multirow{1}{*}{\it MST}   &  MVT+Glove     &  644   & {\bf .719}   & .293     &  687   & {\bf .752}   & .540     &  189   & .732   & {\bf .676}     &  609   & {\bf .586}   & {\bf .282}   &  606   & {\bf .692}   \\ % & .000   %. % \\ \hline \hline 
                                       % &     + FastText    &  000   & .000   & .000     &  000   & .000   & .000     &  000   & .000   & .000     &  000   & .000   & .000    	&  000   & .000  \\ 
                                       % &     + BertEmb    &  000   & .000   & .000     &  000   & .000   & .000     &  000   & .000   & .000     &  000   & .000   & .000    	&  000   & .000   \\
% \hline \hline

%\multicolumn{2}{r|}{{\bf Gain\%} (vs DocNADEe)}    &  000   & {000}   & {000}     &  000   & {000}   & {000}     &  000   & {000}   & 000       &  000   & {000}   & 000    &  000   & {000}

\end{tabular}}
\caption{State-of-the-art comparisons with topic models using word embeddings: PPL,  COH and IR at retrieval fraction $0.02$. 
%KB: Knowledge-base of word and latent topic features. 
Scores are reported on each of the target, given KBs from one or several sources.  %given the word embedding and/or latent topic transfer(s) from one or more related or distant source domains.  
%LVT, GVT and MVT: {\it L}ocal, {\it G}lobal and {\it M}ulti View Transfer; 
%MST: Multi Source Transfer and 
Here, MVT: LVT+GVT (Table \ref{tab:scoresTMwithoutWordEmbeddings}), DocNADEe: DocNADE+Glove. 
%{\it Please read column-wise}. 
%{\bf Bold}: best in column. %\underline{underline}: Scores higher than DocNADEe. 
 %View Transfer, GVT: Topic Transfer.  +Glove: EmbTF+TTF including Glove embeddings. ALL: multi-source transfer. TO DO: Replace EmbTF+TTF with MVT
}
\label{tab:scoresTMwithWordEmbeddings}
\end{table*}

\begin{table*}[t]
\begin{minipage}{.499\linewidth}
\center
%\small
\renewcommand*{\arraystretch}{1.3}
\resizebox{.999\textwidth}{!}{
\setlength\tabcolsep{3.pt}
\begin{tabular}{r|r|ccc|ccc}

\multirow{1}{*}{\bf KBs from}   &     \multirow{1}{*}{\bf Model/} & \multicolumn{6}{c}{{\bf Scores on Target Corpus}}   \\ \cline{3-8}

\multirow{1}{*}{\bf Source}      &   \multirow{1}{*}{\bf Transfer} &   \multicolumn{3}{c|}{\texttt{Ohsumedtitle}}     &  \multicolumn{3}{c}{\texttt{Ohsumed}}   \\
   \multirow{1}{*}{\bf Corpus}   &   \multirow{1}{*}{\bf Type}  & $PPL$  &  $COH$ &  $IR$ &    $PPL$  &  $COH$ &  $IR$   \\ \hline % ($\lambda, \gamma$)  

%\multirow{5}{*}{\it baselines}   &  NVDM    		&  {\bf 951}   & .730   & .085     & {\bf 1310}   & .520   & .105          \\
\multirow{4}{*}{\it baselines}  &  ProdLDA    	&  1121   & .734   & .080     &  1677   & .646   & .080      \\ % say in footnote, DocNADE outperfroms LDA and RSM. ProdLDA outperforms NVDM.
						&   DocNADE     &  1321   & .728   & .160     &  1706   & .662   & .184          \\ %\cdashline{2-8} 

                                               %&  doc2vec         &  000   & .000   & .000     &  000   & .000   & .000          \\
 					       &  EmbSum       &  -   & -   & .150     &  -   & -   & .148          \\   %\cdashline{2-8}
						
						%&  Gauss-LDA     &  000   & .000   & .000     &  000   & .000   & .000          \\    
						%&  glove-DMM    &   000   & .000   & .000     &  000   & .000   & .000          \\                                     
						&  DocNADEe    &   1534   & .738   & .175     &  1637   & .674   &  .183          \\   \hline \hline

\multirow{4}{*}{\it AGnews}    &  LVT     		 &  1587   & .732   & .160    &  1717   & .657   & .184          \\
					  &  GVT     		&  1529   & .732   & .160     &  1594   & .665   & .185          \\
					  &  MVT   			 &  1528   & .734   & .160    &  1598   & .666   & .184          \\
					 &  + BioEmb   		&  1488   & .741   & .176     &  1595   & .676   & .183          \\ \hline

\multirow{4}{*}{\it PubMed}    &  LVT     		 &  {\bf 1268}   & .732   & .172     &  1535   & .669   & .190          \\
					  &  GVT       		&  1392   & .740   & .173     	&  1718   & .671   & {\bf .192}          \\
					  &  MVT    		&  1408   & .743   & .178     	&  1514   & .674   & .191          \\
					 &  + BioEmb     		&  1364   & {\bf .746}   & {\bf .181}     	&  1633   & {\bf .688}   & .183         \\ \hline \hline

\multirow{4}{*}{\it MST}  &  LVT      			&  {\bf 1268}   & .733   & .172     &  1536   & .668   & .190          \\  
					  &  GVT       		&  1391   & .740   & .172     	&  1504   & .666   & {\bf .192}        \\
					  &  MVT    		&  1399   & .743   & .177     	&  1607   & .679   & .191          \\
					 &  + BioEmb    		 &  1375   & .745   & .180    	 &  {\bf 1497}   & .687   & .183        \\ 
                                        % &  + FastText    		 &  0000   & .000   & .000    	 &  000   & .000   & .000        \\
                                       %  &  + BertEmb    		 &  0000   & .000   & .000    	 &  0000   & .000   & .000        \\  \hline \hline

%\multicolumn{2}{r|}{{\bf Gain\%} (vs DocNADE)}        &  000   & 000   & 000     &  000   & 000   & 000    \\
%\multicolumn{2}{r|}{{\bf Gain\%} (vs DocNADEe)}       &  000   & 000   & 000     &  000   & 000   & 000     \\

%\multirow{1}{*}{\bf Gain\%}   &     &  0.0   &  0.0   &  0.0     &  0.0   & 0.0   & 0.0     &  0.0   & 0.0   & 0.0     &  0.0  & 0.0   & 0.0  
\end{tabular}}
\caption{%State-of-the-art comparisons:  
PPL, COH, IR at retrieval fraction $0.02$.  
BioEmb: 200-dimensional word vectors from large biomedical corpus \citep{moen2013distributional}. % of 5+ billion words.
%KB: Knowledge-base of word and latent topic features. 
% Scores are reported on each of the target corpora, given the word embedding and/or latent topic transfer(s) from one or more related or distant source domains.  
%LVT, GVT and MVT: {\it L}ocal, {\it G}lobal and {\it M}ulti View Transfer; 
%MST: Multi Source Transfer and 
+ BioEmb: MVT+BioEmb. 
%Target domain: medical. 
%{\it Please read column-wise}. 
%{\bf Bold}: best in column. %\underline{Underline}: Scores higher than DocNADEe.  
 %View Transfer, GVT: Topic Transfer.  +Glove: EmbTF+TTF including Glove embeddings. ALL: multi-source transfer. TO DO: Replace EmbTF+TTF with MVT
}
\label{tab:scoresMedicaldomain}
\end{minipage} \ \ 
\begin{minipage}{.49\linewidth}
      \center
	\renewcommand*{\arraystretch}{1.3}
	\resizebox{.999\linewidth}{!}{
\setlength\tabcolsep{3.0pt}
\begin{tabular}{c|c|c|l}
%\hline
%\multicolumn{3}{c}{\bf 20NS}                                  \\ \hline
{\bf $\mathcal{T}$}   &     {\bf $\mathcal{S}$}        & \multicolumn{1}{c|}{\bf Model}      & \multicolumn{1}{c}{\bf Topic-words (Top 5)}     \\ \hline
\multirow{6}{*}{\rotatebox{90}{\texttt{20NSshort}}}          &	\multirow{1}{*}{\texttt{20NS}}   	       &  DNE			& {\color{blue} shipping}, sale, prices, {\color{blue} expensive}, price		\\  \cdashline{2-4}  
 							      &			       					    &  {\bf -}GVT			   &	sale, price, monitor, {\color{red} site}, {\color{red} setup}		    \\    
 						             &		   							    &     {\bf +}GVT				&      {\color{blue} shipping}, sale, price, {\color{blue} expensive}, subscribe			\\     \cline{2-4}       

         &	 \multirow{1}{*}{\texttt{AGnews}} 	         &  DNE			&	{\color{blue} microsoft}, {\color{blue} software}, ibm, linux, {\color{blue} computer}		 \\    \cdashline{2-4}
							         &								     	& {\bf -}GVT			&	apple, modem, {\color{red} side}, baud, {\color{red} perform}		   \\    
 						                &					  				&  	{\bf +}GVT		&      {\color{blue} microsoft}, {\color{blue} software}, desktop, {\color{blue} computer}, apple \\	\hline

\multirow{4}{*}{\rotatebox{90}{\texttt{TMNtitle}}}         &		\multirow{1}{*}{\texttt{AGnews}}  	 &  DNE 				&	miners,  earthquake, {\color{blue} explosion},  stormed, quake 		\\   \cdashline{2-4}
						          &		\multirow{1}{*}{\texttt{TMN}}  		 &  DNE 				&	tsunami, quake, japan, earthquake, {\color{blue} radiation} 		\\   \cdashline{2-4}
						           &									        &  {\bf -}GVT			&	{\color{red} strike}, {\color{red} jackson}, kill, earthquake, injures		  \\    
 						             &									&  	{\bf +}GVT					&      earthquake, {\color{blue} radiation}, {\color{blue} explosion}, wildfire  		 
\end{tabular}}
\caption{Source $\mathcal{S}$ and target $\mathcal{T}$ topics before (-) and after (+) topic transfer(s) (GVT) from one or more sources. DNE: DocNADE}
\label{topiccoherenceexamples}
%\end{minipage} 

%\begin{minipage}{.52\linewidth}
      \center
	\renewcommand*{\arraystretch}{1.25}
	\resizebox{.999\linewidth}{!}{
\setlength\tabcolsep{3.pt}
\begin{tabular}{ccc|c|c} \hline
\multicolumn{5}{c}{\bf chip} \\ \hline
\multicolumn{3}{c|}{source corpora} &  \multicolumn{2}{c}{target corpus}         \\ \hline

\multirow{2}{*}{\texttt{20NS}}	&	\multirow{2}{*}{\texttt{R21578}}	&  \multirow{2}{*}{\texttt{AGnews}}		& \multicolumn{2}{c}{\texttt{20NSshort}} \\

	               &						&					&	-GVT	 	&	+GVT  \\  \hline 
key			&		chips 			&	chips 			&	virus	 &	chips 	 		\\
encrypted		&		semiconductor 		&	chipmaker 			&	intel	  &	technology	 \\  
{\color{blue} encryption}		&		miti 				&	 processors 		&	{\color{red} gosh}	 &	intel		 \\  
{\color{blue} clipper}		&		makers 			&	 semiconductor 		&	crash	 &	{\color{blue} encryption}				 \\  
keys			&		semiconductors 		&	intel 				&	chips	   &	{\color{blue} clipper}				 \\   \hline

\end{tabular}}
\caption{Five nearest neighbors of the word {\it chip} in source and target semantic spaces before (-) and after (+) knowledge transfer (MST+GVT)}
\label{tab:NN}
\end{minipage}
\end{table*}

\makeatletter
\def\labelonly{BDF}
\def\labelcheck#1{
    \edef\pgfmarshal{\noexpand\pgfutil@in@{#1}{\labelonly}}
    \pgfmarshal
    \ifpgfutil@in@[#1]\fi
}
\makeatother

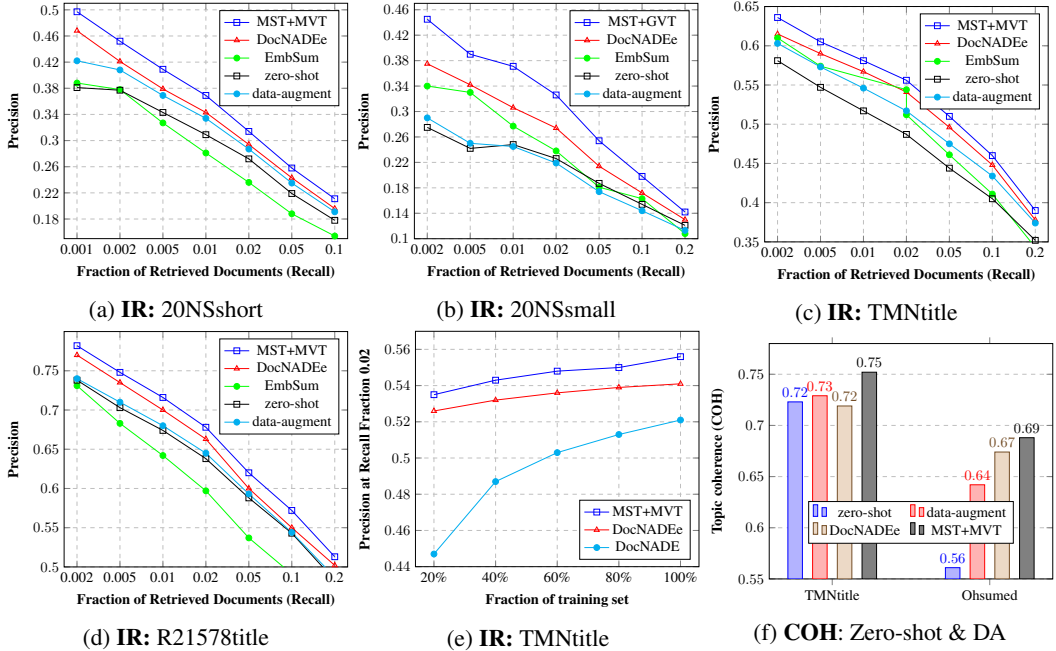
\begin{figure*}[t]
\center
\begin{subfigure}{0.33\textwidth}
\center
\begin{tikzpicture}[scale=0.55][baseline]
\begin{axis}[
    xlabel={\bf Fraction of Retrieved Documents (Recall)},% (\%)},
    ylabel={\bf Precision}, %(\%)},
    xmin=-0.3, xmax=6.3,
    ymin=0.15, ymax=0.51,
   /pgfplots/ytick={.10,.14,...,.51},
    xtick={0,1,2,3,4,5,6},%,8,9, 10,11,12},
    xticklabels={0.001, 0.002, 0.005, 0.01, 0.02, 0.05, 0.1}, % 0.2, 0.3, 0.5},%, 0.8, 1.0},
    %x tick label style={rotate=45,anchor=east},
    legend pos=north east,
    ymajorgrids=true, xmajorgrids=true,
    grid style=dashed,
]

\addplot[
	color=blue,
	mark=square,
	]
	plot coordinates {
    %(0, 0.497)
    (0, 0.497)
    (1, 0.452)
    (2, 0.409)
    (3, 0.369)
    (4, 0.314)
    (5, 0.258)
    (6, 0.211)
    %(7, 0.157)
    %(8, 0.134)
    %(9, 0.102)
    %(11, 0.077)
    %(12, 0.066)
    %(13, 0.00)
	};
\addlegendentry{MST+MVT}

\addplot[
	color=red,
	mark=triangle,
	]
	plot coordinates {
    %(0, 0.468)
    (0, 0.468)
    (1, 0.421)
    (2, 0.379)
    (3, 0.343)
    (4, 0.294)
    (5, 0.243)
    (6, 0.196)
    %(7, 0.151)
    %(8, 0.127)
    %(9, 0.10)
    %(10, 0.077)
    %(11, 0.066)
      %(12, 0.00)
	};
\addlegendentry{DocNADEe}

\iffalse
\addplot[
	color=cyan,
	mark=triangle,
	]
	plot coordinates {
    %(0, 0.454)
    (0, 0.45)
    (1, 0.42)
    (2, 0.38)
    (3, 0.34)
    (4, 0.29)
    (5, 0.24)
    (6, 0.19)
    %(7, 0.151)
    %(8, 0.125)
    %(9, 0.098)
    %(10, 0.075)
    %(11, 0.066)
    %(12, 0.00)
	};
\addlegendentry{DocNADE}
\fi

\addplot[
	color=green,
	mark=*,
	]
	plot coordinates {
    %(0, 0.388)
    (0, 0.388)
    (1, 0.378)
    (2, 0.327)
    (3, 0.281)
    (4, 0.236)
    (5, 0.188)
    (6, 0.154)
    %(7, 0.123)
    %(8, 0.107)
    %(9, 0.088)
    %(10, 0.072)
    %(11, 0.066)
    %(12, 0.00)
	};
\addlegendentry{EmbSum}

\addplot[
	color=black,
	mark=square,
	]
	plot coordinates {
    %(0, 0.497)
    (0, 0.381)
    (1, 0.377)
    (2, 0.343)
    (3, 0.309)
    (4, 0.272)
    (5, 0.219)
    (6, 0.178)
    %(7, 0.138)
    %(8, 0.116)
    %(9, 0.093)
    %(11, 0.074)
    %(12, 0.066)
    %(13, 0.00)
	};
\addlegendentry{zero-shot}

\addplot[
	color=cyan,
	mark=*,
	]
	plot coordinates {
    %(0, 0.497)
    (0, 0.422)
    (1, 0.408)
    (2, 0.369)
    (3, 0.334)
    (4, 0.287)
    (5, 0.235)
    (6, 0.191)
    %(7, 0.146)
    %(8, 0.122)
    %(9, 0.196)
    %(11, 0.175)
    %(12, 0.166)
    %(13, 0.00)
	};
\addlegendentry{data-augment}
\end{axis}
\end{tikzpicture}%
\caption{{\bf IR:} 20NSshort} \label{IR20NSshort}
\end{subfigure}\hspace*{\fill}%
%~~~%
\begin{subfigure}{0.33\textwidth}
\centering
\begin{tikzpicture}[scale=0.55][baseline]
\begin{axis}[
    xlabel={\bf Fraction of Retrieved Documents (Recall)},% (\%)},
    ylabel={\bf Precision},% (\%)},
    xmin=-0.3, xmax=6.3,
    ymin=0.10, ymax=0.47,
   /pgfplots/ytick={.10,.14,...,.47},
    xtick={0,1,2,3,4,5,6},%,8,9, 10,11,12},
    xticklabels={0.002, 0.005, 0.01, 0.02, 0.05, 0.1,  0.2},% 0.3, 0.5},%, 0.8, 1.0},
    %x tick label style={rotate=45,anchor=east},
    legend pos=north east,
    ymajorgrids=true, xmajorgrids=true,
    grid style=dashed,
]

\addplot[
	color=blue,
	mark=square,
	]
	plot coordinates {
    %(0, 0.497)
    (0, 0.445)
    (1, 0.390)
    (2, 0.371)
    (3, 0.326)
    (4, 0.254)
    (5, 0.198)
    (6, 0.142)
    %(7, 0.114)
    %(8, 0.083)
    %(9, 0.060)
    %(10, 0.050)
	};
\addlegendentry{MST+GVT}

\addplot[
	color=red,
	mark=triangle,
	]
	plot coordinates {
    %(0, 0.468)
    (0, 0.375)
    (1, 0.342)
    (2, 0.306)
    (3, 0.274)
    (4, 0.214)
    (5, 0.172)
    (6, 0.130)
    %(7, 0.106)
    %(8, 0.080)
    %(9, 0.059)
    %(10, 0.050)
	};
\addlegendentry{DocNADEe}

\iffalse
\addplot[
	color=cyan,
	mark=triangle,
	]
	plot coordinates {
    %(0, 0.454)
    (0, 0.405)
    (1, 0.367)
    (2, 0.303)
    (3, 0.270)
    (4, 0.209)
    (5, 0.165)
    (6, 0.124)
    %(7, 0.103)
    %(8, 0.079)
    %(9, 0.058)
    %(10, 0.050)
	};
\addlegendentry{DocNADE}
\fi

\addplot[
	color=green,
	mark=*,
	]
	plot coordinates {
    %(0, 0.388)
    (0, 0.340)
    (1, 0.330)
    (2, 0.277)
    (3, 0.238)
    (4, 0.181)
    (5, 0.163)
    (6, 0.108)
    %(7, 0.089)
    %(8, 0.070)
    %(9, 0.056)
    %(10, 0.050)
	};
\addlegendentry{EmbSum}

\addplot[
	color=black,
	mark=square,
	]
	plot coordinates {
    %(0, 0.388)
    (0, 0.275)
    (1, 0.242)
    (2, 0.248)
    (3, 0.226)
    (4, 0.187)
    (5, 0.154)
    (6, 0.121)
    %(7, 0.100)
    %(8, 0.078)
    %(9, 0.058)
    %(10, 0.050)
	};
\addlegendentry{zero-shot}

\addplot[
	color=cyan,
	mark=*,
	]
	plot coordinates {
    %(0, 0.388)
    (0, 0.290)
    (1, 0.250)
    (2, 0.245)
    (3, 0.219)
    (4, .174)
    (5, 0.144)
    (6, 0.113)
    %(7, 0.096)
    %(8, 0.076)
    %(9, 0.058)
    %(10, 0.050)
	};
\addlegendentry{data-augment}

\end{axis}
\end{tikzpicture}%
\caption{{\bf IR:} 20NSsmall} \label{IR20NSsmall}
\end{subfigure}\hspace*{\fill}%
\begin{subfigure}{0.33\textwidth}
\centering
\begin{tikzpicture}[scale=0.55][baseline]
\begin{axis}[
    xlabel={\bf Fraction of Retrieved Documents (Recall)},% (\%)},
    ylabel={\bf Precision},% (\%)},
    xmin=-0.3, xmax=6.3,
    ymin=0.35, ymax=0.65,
   /pgfplots/ytick={.35,.40,...,.65},
    xtick={0,1,2,3,4,5,6},%,8,9, 10,11,12},
    xticklabels={0.002, 0.005, 0.01, 0.02, 0.05, 0.1,  0.2},% 0.3, 0.5},%, 0.8, 1.0},
    %x tick label style={rotate=45,anchor=east},
    legend pos=north east,
    ymajorgrids=true, xmajorgrids=true,
    grid style=dashed,
]

\addplot[
	color=blue,
	mark=square,
	]
	plot coordinates {
    %(0, 0.000)
    (0, 0.636)
    (1, 0.605)
    (2, 0.581)
    (3, 0.556)
    (4, 0.51)
    (5, 0.46)
    (6, 0.39)
    %(7, 0.334)
    %(8, 0.263)
    %(9, 0.199)
    %(10, 0.169)
	};
\addlegendentry{MST+MVT}

\addplot[
	color=red,
	mark=triangle,
	]
	plot coordinates {
    %(0, 0.468)
    (0, 0.615)
    (1, 0.590)
    (2, 0.567)
    (3, 0.541)
    (4, 0.496)
    (5, 0.448)
    (6, 0.378)
    %(7, 0.327)
    %(8, 0.258)
    %(9, 0.197)
    %(10, 0.169)
	};
\addlegendentry{DocNADEe}

\addplot[
	color=green,
	mark=*,
	]
	plot coordinates {
   %(0, 0.694)
    %(1, 0.653)
    %(2, 0.633)
    (0, 0.610)
    (1, 0.574)
    (3, 0.544)
    (3, 0.512)
    (4, 0.461)
    (5, 0.411)
    (6, 0.343)
    (7, 0.296)
    (8, 0.239)
    (9, 0.190)
    (10, 0.169)
	};
\addlegendentry{EmbSum}

\addplot[
	color=black,
	mark=square,
	]
	plot coordinates {
    %(0, 0.468)
    (0, 0.581)
    (1, 0.547)
    (2, 0.517)
    (3, 0.487)
    (4, 0.444)
    (5, 0.405)
    (6, 0.352)
    %(7, 0.312)
    %(8, 0.256)
    %(9, 0.198)
    %(10, 0.169)
	};
\addlegendentry{zero-shot}

\addplot[
	color=cyan,
	mark=*,
	]
	plot coordinates {
    %(0, 0.468)
    (0, 0.603)
    (1, 0.573)
    (2, 0.546)
    (3, 0.517)
    (4, 0.475)
    (5, 0.434)
    (6, 0.374)
    %(7, 0.328)
    %(8, 0.262)
    %(9, 0.199)
    %(10, 0.169)
	};
\addlegendentry{data-augment}

\end{axis}
\end{tikzpicture}%
\caption{{\bf IR:} TMNtitle} \label{IRTMNtitle}
\end{subfigure}\hspace*{\fill}%

\begin{subfigure}{0.33\textwidth}
\centering
\begin{tikzpicture}[scale=0.55][baseline]
\begin{axis}[
    xlabel={\bf Fraction of Retrieved Documents (Recall)},% (\%)},
    ylabel={\bf Precision},% (\%)},
    xmin=-0.3, xmax=6.3,
    ymin=0.50, ymax=0.80,
   /pgfplots/ytick={.5,.55,...,.80},
    xtick={0,1,2,3,4,5,6},%,8,9, 10,11,12},
    xticklabels={0.002, 0.005, 0.01, 0.02, 0.05, 0.1,  0.2},% 0.3, 0.5},%, 0.8, 1.0},
    %x tick label style={rotate=45,anchor=east},
    legend pos=north east,
    ymajorgrids=true, xmajorgrids=true,
    grid style=dashed,
]

\addplot[
	color=blue,
	mark=square,
	]
	plot coordinates {
    %(0, 0.468)
    (0, 0.782)
    (1, 0.748)
    (2, 0.716)
    (3, 0.678)
    (4, 0.620)
    (5, 0.572)
    (6, 0.513)
    %(7, 0.328)
    %(8, 0.262)
    %(9, 0.199)
    %(10, 0.169
	};
\addlegendentry{MST+MVT}

\addplot[
	color=red,
	mark=triangle,
	]
	plot coordinates {
    %(0, 0.468)
    (0, 0.77)
    (1, 0.735)
    (2, 0.70)
    (3, 0.663)
    (4, 0.60)
    (5, 0.55)
    (6, 0.502)
    %(7, 0.328)
    %(8, 0.262)
    %(9, 0.199)
    %(10, 0.169
	};
\addlegendentry{DocNADEe}

\addplot[
	color=green,
	mark=*,
	]
	plot coordinates {
    %(0, 0.468)
    (0, 0.731)
    (1, 0.683)
    (2, 0.642)
    (3, 0.597)
    (4, 0.537)
    (5, 0.491)
    (6, 0.442)
    %(7, 0.328)
    %(8, 0.262)
    %(9, 0.199)
    %(10, 0.169
	};
\addlegendentry{EmbSum}

\addplot[
	color=black,
	mark=square,
	]
	plot coordinates {
    %(0, 0.468)
    (0, 0.738)
    (1, 0.703)
    (2, 0.674)
    (3, 0.638)
    (4, 0.588)
    (5, 0.543)
    (6, 0.483)
    %(7, 0.328)
    %(8, 0.262)
    %(9, 0.199)
    %(10, 0.169
    %(10, 0.169)
	};
\addlegendentry{zero-shot}

\addplot[
	color=cyan,
	mark=*,
	]
	plot coordinates {
    %(0, 0.468)
    (0, 0.740)
    (1, 0.710)
    (2, 0.680)
    (3, 0.645)
    (4, 0.593)
    (5, 0.544)
    (6, 0.487)
    %(7, 0.328)
    %(8, 0.262)
    %(9, 0.199)
    %(10, 0.169)
	};
\addlegendentry{data-augment}
\end{axis}
\end{tikzpicture}%
\caption{{\bf IR:} R21578title} \label{IRR21578title}
\end{subfigure}\hspace*{\fill}%
\begin{subfigure}{0.33\textwidth}
\centering
\begin{tikzpicture}[scale=0.55][baseline]
\begin{axis}[
    xlabel={\bf Fraction of training set},% (\%)},
    ylabel={\bf Precision at Recall Fraction 0.02},
     xmin=-0.3, xmax=4.3,
    ymin=0.44, ymax=0.57,
   /pgfplots/ytick={.44,.46,...,.57},
    xtick={0, 1, 2, 3, 4},
    xticklabels={20\%, 40\%, 60\%, 80\%, 100\%},
    %x tick label style={rotate=45,anchor=east},
    legend pos=south east,
    ymajorgrids=true, xmajorgrids=true,
    grid style=dashed,
]

\addplot[
	color=blue,
	mark=square,
	]
	plot coordinates {
     (0, 0.535)
    (1, 0.543)
    (2, 0.548)
    (3, 0.550)
    (4, 0.556) 
	};
\addlegendentry{MST+MVT}

\addplot[
	color=red,
	mark=triangle,
	]
	plot coordinates {
    (0, 0.526)
    (1, 0.532)
    (2, 0.536)
    (3, 0.539)
    (4, 0.541) 
	};
\addlegendentry{DocNADEe}

\addplot[
	color=cyan,
	mark=*,
	]
	plot coordinates {
    (0, 0.447)
    (1, 0.487)
    (2, 0.503)
    (3, 0.513)
    (4, 0.521) 
	};
\addlegendentry{DocNADE}

\end{axis}
\end{tikzpicture}%
\caption{{\bf IR:} TMNtitle} \label{TMNtitle}
\end{subfigure}\hspace*{\fill}%
\iffalse
\begin{subfigure}{0.33\textwidth}
\centering
\begin{tikzpicture}[scale=0.55][baseline]
\begin{axis}[
    xlabel={\bf Fraction of training set},% (\%)},
    ylabel={\bf macro-F1 score},
     xmin=-0.3, xmax=4.3,
    ymin=0.60, ymax=0.72,
   /pgfplots/ytick={.60,.64,...,.72},
    xtick={0, 1, 2, 3, 4},
    xticklabels={20\%, 40\%, 60\%, 80\%, 100\%},
    %x tick label style={rotate=45,anchor=east},
    legend pos=south east,
    ymajorgrids=true, xmajorgrids=true,
    grid style=dashed,
]

\addplot[
	color=blue,
	mark=square,
	]
	plot coordinates {
     (0, 0.682)
    (1, 0.686)
    (2, 0.689)
    (3, 0.687)
    (4, 0.688) 
	};
\addlegendentry{MST+MVT}

\addplot[
	color=red,
	mark=triangle,
	]
	plot coordinates {
    (0, 0.697)
    (1, 0.697)
    (2, 0.704)
    (3, 0.703)
    (4, 0.700) 
	};
\addlegendentry{DocNADEe}

\addplot[
	color=cyan,
	mark=*,
	]
	plot coordinates {
    (0, 0.618)
    (1, 0.637)
    (2, 0.644)
    (3, 0.650)
    (4, 0.662) 
	};
\addlegendentry{DocNADE}

\end{axis}
\end{tikzpicture}%
\caption{{\bf Classification:} TMNtitle} \label{ClassificationTMNtitle}
\end{subfigure}\hspace*{\fill}%
\fi
\begin{subfigure}{0.33\textwidth}
\centering
\begin{tikzpicture}[scale=0.55][baseline]
\begin{axis}[
    legend pos=north east,
    legend columns=2, 
        legend style={
                    % the /tikz/ prefix is necessary here...
                    % otherwise, it might end-up with `/pgfplots/column 2`
                    % which is not what we want. compare pgfmanual.pdf
            /tikz/column 2/.style={
                column sep=5pt,
            },
        },
    %ybar, 
    ybar=7pt,
    legend style={at={(0.5,0.33)},
    anchor=north,legend columns=2},
    xmin=-0.4, xmax=1.4,
    ymin=0.55, ymax=.780,
    ylabel={\bf Topic coherence (COH)},
    xtick={0, 1},
    xticklabels={TMNtitle, Ohsumed},
    %symbolic x coords={tool8,tool9,tool10},
    %xtick=data,
    nodes near coords,
    nodes near coords align={vertical},
    ymajorgrids=true, %xmajorgrids=true,
    grid style=dashed,
    ]
\addplot coordinates {(0,.723) (1,.561)}; \addlegendentry{zero-shot}
\addplot coordinates {(0,.729) (1,.642)}; \addlegendentry{data-augment}
%\addplot coordinates {(0,.709) (1,.662)}; \addlegendentry{DocNADE}
\addplot coordinates {(0,.719) (1,.674)}; \addlegendentry{DocNADEe}
\addplot coordinates {(0,.752) (1,.688)}; \addlegendentry{MST+MVT}

%\legend{zero-shot,data-augment}
%\legend{target-only,MST+MVT}
\end{axis}
\end{tikzpicture}
\caption{{\bf COH}: Zero-shot \& DA} \label{COH:ZeroshotDataAugTMNtitle}
\end{subfigure}%\hspace*{\fill}%
\caption{%DocNADE Vs iDocNADE: 
(a, b, c, d) Retrieval performance (precision) on 20NSshort, 20NSsmall, TMNtitle and R21578title datasets.  
(e) Precision at recall fraction 0.02, each for a fraction (20\%, 40\%, 60\%, 80\%, 100\%) of the training set of TMNtitle. 
(f) Zero-shot and data-augmentation (DA) experiments for topic coherence on TMNtitle and Ohsumed.} 
%See {\it supplementary}.
\label{fig:docretrieval}
\end{figure*}

\subsection{Generalization: Perplexity (PPL)} 
%{\bf Generalization via Perplexity (PPL)}: 
To evaluate the generative performance in TM, we estimate the log-probabilities for the test documents 
and compute the average held-out perplexity per word as, {\small $PPL$ = $\exp \big( - \frac{1}{N} \sum_{t=1}^{N} \frac{1}{|{\bf v}_t|} \log p({\bf v}_{t}) \big)$},  
where $N$  and $|{\bf v}_t|$ are the number of documents and words in a document ${\bf v}_{t}$, respectively.  

Tables \ref{tab:scoresTMwithoutWordEmbeddings} and \ref{tab:scoresTMwithWordEmbeddings} quantitatively show PPL scores on the five target corpora (four short-text and one long-text) by the baselines
and proposed approaches of knowledge transfers using KBs from one or four sources. 
%knowledge transfers from one or several sources (identical, related or distant).   
%Table \ref{tab:scores} quantitatively shows PPL scores on the four target corpora, each with $H$=$200$ topics determined using the development set. 
Using {\small \texttt{TMN}} (as source) for LVT and MVT on {\small \texttt{TMNtitle}}, we see improved (reduced) PPL scores: ($655$ vs $680$) and ($663$ vs $680$) respectively in comparison to DocNADEe.
Similarly using {\small \texttt{AGnews}} (as source) for LVT on TMN target,  we observe improved scores:  ($564$ vs $584$) and  ($564$ vs $572$) compared to DocNADE and DocNADEe, respectively. 
It suggests a positive knowledge transfer and verifies domain relatedness in {\small \texttt{TMN}}-{\small \texttt{TMNtitle}} and {\small \texttt{AGnews}}-{\small \texttt{TMN}} (Table \ref{tab:domainoverlap}). 
Additionally, we also observe better generalization by MST+LVT on several target corpora, e.g., on {\small \texttt{TMNtitle}}: ($663$ vs $706$) and ($663$ vs $680$) compared to DocNADE and DocNADEe, respectively. 
%Also, MST+LVT boosts ($663$ vs $680$) generalization in {\small \texttt{TMNtitle}}.  
% baselines NVDM and prolda performs poorly
% GVT only does not help in generalization..compare to LVT... 

In Table \ref{tab:scoresMedicaldomain}, we demonstrate generalization performance via PPL on two medical target corpora: {\small \texttt{Ohsumtitle}}  and {\small \texttt{Ohsumed}} by 
knowledge transfer from {\small \texttt{AGnews}}  ({\it news corpus}) and {\small \texttt{PubMed}} ({\it medical} abstracts). 
We see that using {\small \texttt{PubMed}} for LVT on both the target corpora improves generalization: ($1268$ vs $1534$) and ($1535$ vs $1637$) compared to DocNADEe, respectively.
Additionally, MST+GVT and MST+MVT lead to better generalization, compared to DocNADEe.

\subsection{Interpretabilty: Topic Coherence (COH)}
%{\bf Interpretabilty via Topic Coherence (COH)}: 
Beyond perplexity, we compute topic coherence to estimate the meaningfulness of words in each of the topics captured. 
In doing so, we choose the coherence measure proposed by \citet{DBLP:conf/wsdm/RoderBH15}   %\citepauthor{Michael:82}  \shortcite{Michael:82}.  
that identifies context features for each topic word using a sliding 
window over the reference corpus. 
%To estimate meaningfulness of words in the topics captured, 
We follow \citet{pankajgupta:2019iDocNADEe} and compute COH with the top 10 words in each topic. %of the top 10 words in each of the topics 
%We choose the coherence measure proposed by \citept{DBLP:conf/wsdm/RoderBH15}   %\citepauthor{Michael:82}  \shortcite{Michael:82}.  
%that identifies context features for each topic word using a sliding 
%window over the reference corpus. 
Essentially, higher scores imply more coherent topics. 
%to assess the meaningfulness of the %captured 
%underlying topics captured. 
%We choose the coherence measure proposed by \citept{Michael:82}   %\citepauthor{Michael:82}  \shortcite{Michael:82}.  
%, which identifies context features for each topic word using a sliding 
%window over the reference corpus. Higher scores imply more coherent topics.

Tables \ref{tab:scoresTMwithoutWordEmbeddings} and \ref{tab:scoresTMwithWordEmbeddings} (under {COH} column) demonstrate  that %average coherence, where  %on the target corpora in sparse-data setting. % using short-text and long-text datasets,  
our proposed knowledge transfer approaches show noticeable gains in COH, e.g., 
using {\small \texttt{AGnews}} as a source alone in GVT configuration for {\small \texttt{20NSsmall}} datatset,  
we observe COH of ($.563$ vs $.455$) compared to DocNADEe. In MVT+Glove and MST+MVT, it is increased to $.573$ and $.600$, respectively. 
Importantly, we find MVT$>$GVT$>$LVT in COH scores for both the single-source and multi-source transfers. 
Here, MST+MVT boosts {COH} for all the five target corpora compared to the baseline (i.e., DocNADE and DocNADEe) topic models. This suggests that there is a  
need for the two complementary (word and topics) representations % for meaningful learning in $\mathcal{T}$. 
and knowledge transfers from several domains in order to guide meaningful learning in $\mathcal{T}$. 
Table \ref{tab:scoresMedicaldomain} also shows similar gains in COH due to GVT on {\small \texttt{Ohsumedtitle}} and   {\small \texttt{Ohsumed}}, 
using latent knowledge from {\small \texttt{PubMed}}.  The results on both the low- and high-resource targets conclude that the proposed modeling scales.

\subsection{Applicability: Information Retrieval (IR)}

To evaluate document representations, we perform a document retrieval task 
on the target datasets and use their label information to compute precision.  
We follow the experimental setup similar to \citet{DBLP:journals/jmlr/LaulyZAL17}, 
where all test documents are treated as queries to retrieve a fraction of the closest documents in
the original training set using cosine similarity  between their document vectors.  
To compute retrieval precision for each fraction (e.g., $0.02$),  %(e.g., $0.001$, $0.005$, etc.), 
we average the number of retrieved training documents with the same label as the query. 
%Since,  \citept{Salakhutdinov:82}  and  \citept{HugoJMLR:82}  have shown that RSM and DocNADE strictly outperform LDA on this task, 
%we solely compare DocNADE  with our proposed extensions.

Tables \ref{tab:scoresTMwithoutWordEmbeddings} and \ref{tab:scoresTMwithWordEmbeddings} depict precision scores at retrieval fraction $0.02$ (similar to \citet{pankajgupta:2019iDocNADEe}),  
where the configuration MST+MVT outperforms both the DocNADE and 
DocNADEe in retrieval performance 
on the four target (short-text)  datasets, e.g., ($.556$ vs $.521$) and ($.556$ vs $.541$)
for {\small \texttt{TMNtitle}}, respectively. 
A gain in IR performance is noticeable for highly overlapping domains, e.g., {\small \texttt{TMN}}-{\small \texttt{TMNtitle}} 
than the related, e.g., {\small \texttt{AGnews}}-{\small \texttt{TMNtitle}}. 
We also see a large gain ($.326$ vs $.270$) in DocNADE due to MST+GVT for {\small \texttt{20NSsmall}}. 
Similarly, Table \ref{tab:scoresMedicaldomain} shows improved precision on medical corpora, where    
MVT+BioEmb and GVT using {\small \texttt{PubMed}} report gains ($.181$ vs $.160$ and $.192$ vs $.184$) on {\small \texttt{Ohsumedtitle}}  and {\small \texttt{Ohsumed}}, respectively.   
Additionally, Figures \ref{IR20NSshort}, \ref{IR20NSsmall}, \ref{IRTMNtitle} and \ref{IRR21578title} illustrate the precision on  
{\small \texttt{20NSshort}}, {\small \texttt{20NSsmall}}, {\small \texttt{TMNtitle}} and {\small \texttt{R21578title}}, respectively, 
where the proposed approaches (MST+GVT and MST+MVT) consistently outperform the baselines at all fractions. 
The IR results on both the low- and high-resource targets imply that our approaches scale.  

Moreover, we split the training data of TMNtitle into several sets: 20\%, 40\%, 60\%, 80\% of the training set 
and then retrain DocNADE, DocNADEe and DocNADE+MST+MVT. 
We demonstrate the impact of knowledge transfers via word and topic 
features in learning representations on the sparse target domain. 
Figure \ref{TMNtitle} plots precision at retrieval (recall) fraction $0.02$ and 
demonstrates that the proposed modeling consistently reports a gain over DocNADE(e) at each of the splits.

%\subsection{Applicability: Text Classification}
\subsection{Zero-shot and Data-augmentation Evaluations} 

Figures \ref{IR20NSshort}, \ref{IR20NSsmall}, \ref{IRTMNtitle} and \ref{IRR21578title} show precision in the zero-shot (source-only training) and data-augmentation (source+target training) configurations. 
Observe that the latter helps in learning meaningful representations and performs better than zero-shot; however, 
it is  outperformed by MST+MVT, suggesting that a naive (data space) augmentation does not add sufficient prior or relevant information to the sparse target.    
Thus, we find that it is beneficial to augment training data in feature space (e.g., LVT, GVT and MVT) especially for unsupervised topic models using latent knowledge from one or several relevant sources. 

Beyond IR, we further investigate computing topic coherence (COH) for zero-shot and data-augmentation baselines, where 
the COH scores (Figure \ref{COH:ZeroshotDataAugTMNtitle}) suggest that MST+MVT outperforms DocNADEe, zero-shot and data-augmentation. 

%(To Do: Include a a detailed explanation)

%\subsection{Qualitative Analysis: Source-Target Agreements for Word and Topic Representations} 
\subsection{Qualitative Analysis: Topics and Nearest Neighbors (NN)} 
For topic level inspection, we first extract topics using the rows of ${\bf W}$ of source and target corpora. Table \ref{topiccoherenceexamples}  
demonstrates that topics in the target domains become more coherent due to GVT(+MST).   Observe that we also show topics from source domain(s) that align with the topics from target. 

For word level inspection, we extract word representations using the columns of ${\bf W}$. 
Table \ref{tab:NN} shows nearest neighbors (NNs) of the word {\it chip} in {\small \texttt{20NSshort}} (target) corpus, before and after GVT using three knowledge sources.
Observe that the NNs in the target become more meaningful. 

%(To Do: Include a a detailed explanation)

%using the columns W:,vi  

%- topic example with source-target topic alignment, how to get the topics

%To analyze the
%meaningful semantics captured, we perform a qualitative inspection
%of the learned representations by the topic models.
%Table 5 shows topics for 20NS dataset that could be interpreted
%as religion, which are (sub)categories in the data,
%confirming that meaningful topics are captured. Observe that
%DocNADEe extracts a more coherent topic.
%For word level inspection, we extract word representations
%using the columns W:,vi as the vector (200 dimension)
%representation of each word vi, learned by iDocNADE using
%20NS dataset. Table 6 shows the five nearest neighbors
%of some selected words in this space and their corresponding
%similarity scores. We also compare similarity in word
%vectors from iDocNADE and glove embeddings, confirming
%that meaningful word representations are learned.

% Qualitatively, 
%Table \ref{topiccoherenceexamples} illustrates example topics from target domains, 
%where GVT using a corresponding source shows more coherent topics.  
%the 20NSshort text dataset %with their coherence scores 
%for DocNADE,  ctx-DocNADE  and ctx-DocNADEe, where the inclusion of embeddings results in a more coherent topic.  

%-  nearest neighbours, how to get them 
%- source-target word NN, glove also, with relevance scores  
%- IR retrieval examples 

\section{Conclusion}
Within neural topic modeling, we have presented approaches to introduce (external) complementary knowledge: pre-trained word embeddings (i.e., local semantics)
and latent topics (i.e., global semantics) exclusively or jointly from one or many sources (i.e., multi-view and multi-source) %in neural topic modeling 
that better deal with data-sparsity issues, especially in a short-text and/or small document collection.  
We have shown learning meaningful topics and text representations on 7 (low- and high-resource) target corpora from news and medical domains.  
%that we have evaluated via perplexity, topic coherence and retrieval accuracy using knowledge from 6 (large) sources.%on  five target corpora from news and medical domains. 

\bibliography{iclr2020_conference}
\bibliographystyle{iclr2020_conference}

\end{document}

%% file: math_commands.tex
\usepackage{amsmath,amsfonts,bm}

\def\eqref#1{equation~\ref{#1}}
\def\1{\bm{1}}

\DeclareMathAlphabet{\mathsfit}{\encodingdefault}{\sfdefault}{m}{sl}
\SetMathAlphabet{\mathsfit}{bold}{\encodingdefault}{\sfdefault}{bx}{n}

%% file: iclr2020_conference.bbl
\begin{thebibliography}{18}
\providecommand{\natexlab}[1]{#1}
\providecommand{\url}[1]{\texttt{#1}}
\expandafter\ifx\csname urlstyle\endcsname\relax
  \providecommand{\doi}[1]{doi: #1}\else
  \providecommand{\doi}{doi: \begingroup \urlstyle{rm}\Url}\fi

\bibitem[Bengio et~al.(2003)Bengio, Ducharme, Vincent, and
  Janvin]{DBLP:journals/jmlr/BengioDVJ03}
Yoshua Bengio, R{\'{e}}jean Ducharme, Pascal Vincent, and Christian Janvin.
\newblock A neural probabilistic language model.
\newblock \emph{Journal of Machine Learning Research}, 3:\penalty0 1137--1155,
  2003.
\newblock URL \url{http://www.jmlr.org/papers/v3/bengio03a.html}.

\bibitem[Blei et~al.(2003)Blei, Ng, and Jordan]{DBLP:journals/jmlr/BleiNJ03}
David~M. Blei, Andrew~Y. Ng, and Michael~I. Jordan.
\newblock Latent dirichlet allocation.
\newblock \emph{Journal of Machine Learning Research}, 3:\penalty0 993--1022,
  2003.
\newblock URL \url{http://www.jmlr.org/papers/v3/blei03a.html}.

\bibitem[Cao et~al.(2010)Cao, Pan, Zhang, Yeung, and
  Yang]{DBLP:conf/aaai/CaoPZYY10}
Bin Cao, Sinno~Jialin Pan, Yu~Zhang, Dit{-}Yan Yeung, and Qiang Yang.
\newblock Adaptive transfer learning.
\newblock In \emph{Proceedings of the Twenty-Fourth {AAAI} Conference on
  Artificial Intelligence, {AAAI} 2010, Atlanta, Georgia, USA, July 11-15,
  2010}. {AAAI} Press, 2010.
\newblock URL
  \url{http://www.aaai.org/ocs/index.php/AAAI/AAAI10/paper/view/1823}.

\bibitem[Das et~al.(2015)Das, Zaheer, and Dyer]{P15-1077}
Rajarshi Das, Manzil Zaheer, and Chris Dyer.
\newblock Gaussian lda for topic models with word embeddings.
\newblock In \emph{Proceedings of the 53rd Annual Meeting of the Association
  for Computational Linguistics and the 7th International Joint Conference on
  Natural Language Processing (Volume 1: Long Papers)}, pp.\  795--804.
  Association for Computational Linguistics, 2015.
\newblock \doi{10.3115/v1/P15-1077}.
\newblock URL \url{http://aclweb.org/anthology/P15-1077}.

\bibitem[Gupta et~al.(2019)Gupta, Chaudhary, Buettner, and
  Sch{\"u}tze]{pankajgupta:2019iDocNADEe}
Pankaj Gupta, Yatin Chaudhary, Florian Buettner, and Hinrich Sch{\"u}tze.
\newblock Document informed neural autoregressive topic models with
  distributional prior.
\newblock In \emph{Proceedings of the Thirty-Third AAAI Conference on
  Artificial Intelligence}, 2019.
\newblock URL \url{{http://arxiv.org/abs/1809.06709}}.

\bibitem[Larochelle \& Lauly(2012)Larochelle and
  Lauly]{DBLP:conf/nips/LarochelleL12}
Hugo Larochelle and Stanislas Lauly.
\newblock A neural autoregressive topic model.
\newblock In Peter~L. Bartlett, Fernando C.~N. Pereira, Christopher J.~C.
  Burges, L{\'{e}}on Bottou, and Kilian~Q. Weinberger (eds.), \emph{Advances in
  Neural Information Processing Systems 25: 26th Annual Conference on Neural
  Information Processing Systems}, pp.\  2717--2725, 2012.
\newblock URL
  \url{http://papers.nips.cc/paper/4613-a-neural-autoregressive-topic-model}.

\bibitem[Larochelle \& Murray(2011)Larochelle and
  Murray]{DBLP:journals/jmlr/LarochelleM11}
Hugo Larochelle and Iain Murray.
\newblock The neural autoregressive distribution estimator.
\newblock In Geoffrey~J. Gordon, David~B. Dunson, and Miroslav Dud{\'{\i}}k
  (eds.), \emph{Proceedings of the Fourteenth International Conference on
  Artificial Intelligence and Statistics, AISTATS}, volume~15 of \emph{{JMLR}
  Proceedings}, pp.\  29--37. JMLR.org, 2011.

\bibitem[Lauly et~al.(2017)Lauly, Zheng, Allauzen, and
  Larochelle]{DBLP:journals/jmlr/LaulyZAL17}
Stanislas Lauly, Yin Zheng, Alexandre Allauzen, and Hugo Larochelle.
\newblock Document neural autoregressive distribution estimation.
\newblock \emph{Journal of Machine Learning Research}, 18:\penalty0
  113:1--113:24, 2017.
\newblock URL \url{http://jmlr.org/papers/v18/16-017.html}.

\bibitem[Le \& Mikolov(2014)Le and Mikolov]{DBLP:conf/icml/LeM14}
Quoc~V. Le and Tomas Mikolov.
\newblock Distributed representations of sentences and documents.
\newblock In \emph{Proceedings of the 31th International Conference on Machine
  Learning, {ICML}}, volume~32 of \emph{{JMLR} Workshop and Conference
  Proceedings}, pp.\  1188--1196. JMLR.org, 2014.
\newblock URL \url{http://jmlr.org/proceedings/papers/v32/le14.html}.

\bibitem[Miao et~al.(2016)Miao, Yu, and Blunsom]{DBLP:conf/icml/MiaoYB16}
Yishu Miao, Lei Yu, and Phil Blunsom.
\newblock Neural variational inference for text processing.
\newblock In \emph{Proceedings of the 33nd International Conference on Machine
  Learning, {ICML} 2016, New York City, NY, USA, June 19-24, 2016}, volume~48
  of \emph{{JMLR} Workshop and Conference Proceedings}, pp.\  1727--1736.
  JMLR.org, 2016.

\bibitem[Mikolov et~al.(2013)Mikolov, Sutskever, Chen, Corrado, and
  Dean]{DBLP:conf/nips/MikolovSCCD13}
Tomas Mikolov, Ilya Sutskever, Kai Chen, Gregory~S. Corrado, and Jeffrey Dean.
\newblock Distributed representations of words and phrases and their
  compositionality.
\newblock In Christopher J.~C. Burges, L{\'{e}}on Bottou, Zoubin Ghahramani,
  and Kilian~Q. Weinberger (eds.), \emph{Advances in Neural Information
  Processing Systems 26: 27th Annual Conference on Neural Information
  Processing Systems}, pp.\  3111--3119, 2013.
\newblock URL
  \url{http://papers.nips.cc/paper/5021-distributed-representations-of-words-and-phrases-and-their-compositionality}.

\bibitem[Moen \& Ananiadou(2013)Moen and Ananiadou]{moen2013distributional}
SPFGH Moen and Tapio Salakoski2~Sophia Ananiadou.
\newblock Distributional semantics resources for biomedical text processing.
\newblock \emph{Proceedings of LBM}, pp.\  39--44, 2013.

\bibitem[Nguyen et~al.(2015)Nguyen, Billingsley, Du, and
  Johnson]{DBLP:journals/tacl/NguyenBDJ15}
Dat~Quoc Nguyen, Richard Billingsley, Lan Du, and Mark Johnson.
\newblock Improving topic models with latent feature word representations.
\newblock \emph{{TACL}}, 3:\penalty0 299--313, 2015.
\newblock URL
  \url{https://tacl2013.cs.columbia.edu/ojs/index.php/tacl/article/view/582}.

\bibitem[Pennington et~al.(2014)Pennington, Socher, and Manning]{D14-1162}
Jeffrey Pennington, Richard Socher, and Christopher Manning.
\newblock Glove: Global vectors for word representation.
\newblock In \emph{Proceedings of the 2014 Conference on Empirical Methods in
  Natural Language Processing (EMNLP)}, pp.\  1532--1543. Association for
  Computational Linguistics, 2014.
\newblock \doi{10.3115/v1/D14-1162}.
\newblock URL \url{http://aclweb.org/anthology/D14-1162}.

\bibitem[Peters et~al.(2018)Peters, Neumann, Iyyer, Gardner, Clark, Lee, and
  Zettlemoyer]{N18-1202}
Matthew Peters, Mark Neumann, Mohit Iyyer, Matt Gardner, Christopher Clark,
  Kenton Lee, and Luke Zettlemoyer.
\newblock Deep contextualized word representations.
\newblock In \emph{Proceedings of the 2018 Conference of the North American
  Chapter of the Association for Computational Linguistics: Human Language
  Technologies, Volume 1 (Long Papers)}, pp.\  2227--2237. Association for
  Computational Linguistics, 2018.
\newblock \doi{10.18653/v1/N18-1202}.
\newblock URL \url{http://aclweb.org/anthology/N18-1202}.

\bibitem[R{\"{o}}der et~al.(2015)R{\"{o}}der, Both, and
  Hinneburg]{DBLP:conf/wsdm/RoderBH15}
Michael R{\"{o}}der, Andreas Both, and Alexander Hinneburg.
\newblock Exploring the space of topic coherence measures.
\newblock In \emph{Proceedings of the Eighth {ACM} International Conference on
  Web Search and Data Mining, {WSDM} 2015, Shanghai, China, February 2-6,
  2015}, pp.\  399--408. {ACM}, 2015.
\newblock URL \url{https://doi.org/10.1145/2684822.2685324}.

\bibitem[Salakhutdinov \& Hinton(2009)Salakhutdinov and
  Hinton]{DBLP:conf/nips/SalakhutdinovH09}
Ruslan Salakhutdinov and Geoffrey~E. Hinton.
\newblock Replicated softmax: an undirected topic model.
\newblock In Yoshua Bengio, Dale Schuurmans, John~D. Lafferty, Christopher
  K.~I. Williams, and Aron Culotta (eds.), \emph{Advances in Neural Information
  Processing Systems 22: 23rd Annual Conference on Neural Information
  Processing Systems}, pp.\  1607--1614. Curran Associates, Inc., 2009.
\newblock URL
  \url{http://papers.nips.cc/paper/3856-replicated-softmax-an-undirected-topic-model}.

\bibitem[Srivastava \& Sutton(2017)Srivastava and Sutton]{SrivastavaSutton}
Akash Srivastava and Charles Sutton.
\newblock Autoencoding variational inference for topic models.
\newblock In \emph{5th International Conference on Learning Representations,
  ICLR}, 2017.
\newblock URL \url{https://arxiv.org/pdf/1703.01488.pdf}.

\end{thebibliography}
